\documentclass[10pt]{iopart}

\usepackage{cite}
\usepackage{fancyhdr}

\usepackage{graphicx}
\usepackage{setspace}
\usepackage{epstopdf}
\usepackage{trimclip}

\usepackage{subfiles}

\usepackage{iopams}
\expandafter\let\csname equation*\endcsname\relax
\expandafter\let\csname endequation*\endcsname\relax
\usepackage{amsmath}

\usepackage[font=small]{caption}
\usepackage{subfig}
\usepackage{amssymb}
\usepackage[export]{adjustbox}

\usepackage{color, soul}
\makeatletter
\AtBeginDocument{\let\hl\@firstofone}
\makeatother

\usepackage{url}

\usepackage{algorithm}
\usepackage{etoolbox}\AtBeginEnvironment{algorithmic}{\scriptsize}
\usepackage[noend]{algpseudocode}

\usepackage{booktabs}

\usepackage{listings}

\usepackage{numprint}

\usepackage{siunitx}
\usepackage[nolist]{acronym}

\usepackage{breqn}

\usepackage[utf8]{inputenc}

\usepackage{todonotes}

\usepackage[shortcuts]{extdash}

\usepackage{textcomp}

\makeatletter
\newcommand\footnoteref[1]{\protected@xdef\@thefnmark{\ref{#1}}\@footnotemark}
\makeatother

\makeatletter
\def\lst@makecaption{%
  \def\@captype{table}%
  \@makecaption
}
\makeatother

\makeatletter
\def\BState{\State\hskip-\ALG@thistlm}
\makeatother

\usepackage[unicode]{hyperref}
\hypersetup{
  colorlinks,
  linkcolor=black,
}

\newcommand{\vect}[1]{\boldsymbol{#1}}

\newcommand{\mset}[1]{\mathcal{#1}}
\newcommand{\edist}[1]{\|#1\|_{2}}

\newcommand{\threeD}{\mbox{3D} }
\newcommand{\twoD}{\mbox{2D} }

\usepackage{tikz}
\usepackage{pgfplots}
\usepackage{filecontents}
\usetikzlibrary{shapes,fit,arrows,calc,intersections,positioning,backgrounds,decorations.markings,patterns,external}
\pgfplotsset{compat=newest}
\pgfdeclarelayer{background}
\pgfdeclarelayer{foreground}
\pgfsetlayers{background,main,foreground}

\tikzset{external/system call={latex \tikzexternalcheckshellescape -halt-on-error
    -interaction=batchmode -jobname "\image" "\texsource";
    dvips -o "\image".eps "\image".dvi;
ps2eps "\image.eps"}}

\tikzset{
  connect/.style args={(#1) to (#2) over (#3) by #4}{
    insert path={
      let \p1=($(#1)-(#3)$), \n1={veclen(\x1,\y1)},
      \n2={atan2(\x1,\y1)}, \n3={abs(#4)}, \n4={#4>0 ?180:-180}  in
      (#1) -- ($(#1)!\n1-\n3!(#3)$)
      arc (\n2:\n2+\n4:\n3) -- (#2)
    }
  },
}

\tikzset{
  state/.style={
    rectangle,
    draw=black, very thick,
    minimum height=1.0em,
    text centered,
  },
  final_state/.style={
    rectangle,
    rounded corners,
    draw=black, very thick,
    minimum height=2em,
    text centered,
  },
  initial_state/.style={
    rectangle,
    double=white,
    double distance=1pt,
    inner sep=2pt,
    draw=black, very thick,
    minimum height=2em,
    text centered,
  },
  point/.style={
    circle,
    inner sep=0pt,
    minimum size=3pt,
    fill=red
  },
  adder/.style={
    circle,
    inner sep=2pt,
    minimum size=0.3in,
    draw=black, very thick,
    text centered
  }
}

\usepackage{enumerate}

\tikzset{
  connect/.style args={(#1) to (#2) over (#3) by #4}{
    insert path={
      let \p1=($(#1)-(#3)$), \n1={veclen(\x1,\y1)},
      \n2={atan2(\x1,\y1)}, \n3={abs(#4)}, \n4={#4>0 ?180:-180}  in
      (#1) -- ($(#1)!\n1-\n3!(#3)$)
      arc (\n2:\n2+\n4:\n3) -- (#2)
    }
  },
}

\usepackage{scalerel}
\usetikzlibrary{svg.path}
\definecolor{orcidlogocol}{HTML}{A6CE39}
\tikzset{
  orcidlogo/.pic={
    \fill[orcidlogocol] svg{M256,128c0,70.7-57.3,128-128,128C57.3,256,0,198.7,0,128C0,57.3,57.3,0,128,0C198.7,0,256,57.3,256,128z};
    \fill[white] svg{M86.3,186.2H70.9V79.1h15.4v48.4V186.2z}
    svg{M108.9,79.1h41.6c39.6,0,57,28.3,57,53.6c0,27.5-21.5,53.6-56.8,53.6h-41.8V79.1z M124.3,172.4h24.5c34.9,0,42.9-26.5,42.9-39.7c0-21.5-13.7-39.7-43.7-39.7h-23.7V172.4z}
    svg{M88.7,56.8c0,5.5-4.5,10.1-10.1,10.1c-5.6,0-10.1-4.6-10.1-10.1c0-5.6,4.5-10.1,10.1-10.1C84.2,46.7,88.7,51.3,88.7,56.8z};
  }
}
\newcommand\orcidicon[1]{\href{https://orcid.org/#1}{\mbox{\scalerel*{
        \begin{tikzpicture}[yscale=-1,transform shape]
          \pic{orcidlogo};
        \end{tikzpicture}
}{|}}}}

\begin{document}

\newcommand{\PREPRINTYEAR}{2020}
\newcommand{\PREPRINTPUBLISHER}{IOP Publishing}
\newcommand{\DOI}{10.1088/1748-3190/abc6b3}

\markboth{\PREPRINTPUBLISHER, \PREPRINTYEAR. DOI: \DOI}{}

\fancyhead{}
\chead{©\PREPRINTPUBLISHER, \PREPRINTYEAR. DOI: \DOI}
\pagestyle{fancy}
\thispagestyle{plain}

\onecolumn
\pagenumbering{gobble}
{
  \topskip0pt
  \vspace*{\fill}
  \centering
  \LARGE{%
    © \PREPRINTYEAR~\PREPRINTPUBLISHER\\\vspace{1cm}
    Personal use of this material is permitted.
    Permission from \PREPRINTPUBLISHER~must be obtained for all other uses, in any current or future media, including reprinting or republishing this material for advertising or promotional purposes, creating new collective works, for resale or redistribution to servers or lists, or reuse of any copyrighted component of this work in other works.\\\vspace*{1cm}DOI: \DOI}
    \vspace*{\fill}

}
\newpage
\pagenumbering{arabic}

\title[Swarms of Unmanned Aerial Vehicles without Communication and External Localization]{Bio-Inspired Compact Swarms of Unmanned Aerial Vehicles without Communication and External Localization}

\author{
  Pavel Petr\'{a}\v{c}ek$^{1\orcidicon{0000-0002-0887-9430}}$,
  Viktor Walter$^{\orcidicon{0000-0001-8693-6261}}$,
  \hl{Tom\'{a}\v{s} B\'{a}\v{c}a}$^{\orcidicon{0000-0001-9649-8277}}$,
  Martin Saska$^{\orcidicon{0000-0001-7106-3816}}$
}

\address{Department of Cybernetics, Faculty of Electrical Engineering, Czech Technical University in Prague, 166 36, Prague 6, Czech Republic}
\address{$^1$Author to whom any correspondence should be addressed.}
\eads{\mailto{pavel.petracek@fel.cvut.cz}, \mailto{viktor.walter@fel.cvut.cz}, \mailto{tomas.baca@fel.cvut.cz}, \mailto{martin.saska@fel.cvut.cz}}
\vspace{10pt}
\begin{indented}
\item[]August 2020
\end{indented}



\begin{abstract}


  This article presents a unique framework for deploying decentralized and infrastructure-independent swarms of homogeneous aerial vehicles in the real world without explicit communication.
  This is a requirement in swarm research, which anticipates that global knowledge and communication will not scale well with the number of robots.
  The system architecture proposed in this article employs the \ac{UVDAR} technique to directly perceive the local neighborhood for direct mutual localization of swarm members.
  The technique allows for decentralization and high scalability \hl{of swarm systems}, such as can be observed in fish schools, bird flocks\hl{,} or cattle herds.
  \hl{%
    The bio-inspired swarming model that has been developed is suited for real-world deployment of large particle groups in outdoor and indoor environments with obstacles.
    The collective behavior of the model emerges from a set of local rules based on direct observation of the neighborhood using onboard sensors only.
    The model is scalable, requires only local perception of agents and the environment, and requires no communication among the agents.
    Apart from simulated scenarios, the performance and usability of the entire framework is analyzed in several real-world experiments with a fully-decentralized swarm of} \acp{UAV} \hl{deployed in outdoor conditions.
  }
  To the best of our knowledge, these experiments are the first deployment of decentralized bio-inspired compact swarms of \acp{UAV} without the use of a communication network or shared absolute localization.
  The entire system is available as open\hl{-}source at \url{https://github.com/ctu-mrs}.
\end{abstract}

\vspace{2pc}

\submitto{\BB}

\maketitle
 
\ioptwocol


\section{Introduction}
\label{sec:introduction}


Use of a team instead of a single robot may yield several general advantages in tasks that either benefit from the multi-robot configuration or are altogether unsolvable by a single robot.
The main advantages of robot teams are reduced task execution time, improved robustness, redundancy, fault tolerance, and convenience of cooperative abilities, such as increased precision of measurements with a stochastic element (e.g., localizing ionizing radiation sources~\cite{9024023}), distributing the application payload, and dynamic collaboration (e.g., cooperative object transport~\cite{7989609}).

Deployment of a single \ac{UAV} requires a complex system composed of several intricate subsystems handling the vehicle control, environment perception, absolute or relative localization, mapping, navigation, and communication.
A system scaled to a set of tightly cooperating \acp{UAV} must additionally introduce decentralized behavior generation, fault detection, information sharing in an often low-to-none bandwidth communication network, and detection and localization of inter-swarm members.
Furthermore, the characteristic environments in the context of aerial swarms suited for real-world challenges may be unknown in advance, they incorporate high density of complex obstacles, they provide none-to-low access to mutual intercommunication between the team agents, and they allow either no access or unreliable access to a \ac{GNSS}.
Each of these concepts is a complex challenge on its own.
However, overcoming all the challenges opens the way to applications requiring distributed sensing and acting, such as cooperative area coverage for search \& rescue, exploration, or surveillance tasks.

In this article, we present a complete swarm system framework, which respects the swarm and environment characteristics.
The properties of the framework presented here correspond closely with the definition of autonomous swarms, as listed in~\cite{Trianni2008EvolutionarySR}.
The properties are: scalability for large groups, high redundancy and fault tolerance, usability in tasks unsolvable by a single robot, and locally limited sensing and communication abilities.
Inspired by the self-organizing behavior of large swarms of homogeneous units with limited local information that is found among biological systems, our framework goes even further beyond the swarm requirements from~\cite{Trianni2008EvolutionarySR} by dealing with all centralized and decentralized communication with the use of the \ac{UVDAR} local perception method.
\hl{%
The elimination of communication is particularly important in dense swarms of fast-moving aerial vehicles, where time-based delays in mutual localization might disturb the collective behavior of swarms and thus may induce mutual collisions.
The independence from communication makes the system also applicable as a backup solution for swarm stabilization in scenarios where communication is required, but suffers from outages.%
}

This allows us to employ a fully decentralized system architecture not limited by scalability constraints.
This decentralization is advantageously robust towards a single-point of failure, reduces the hardware demands for individuals, and distributes the sensing and acting properties.
We have been inspired mainly by ordinary representatives of biological systems: common starlings \textit{sturnus vulgaris}, which exhibit a remarkable ability to maintain cohesion as a group in highly uncertain environments and with limited, noisy information~\cite{sturnus_vulgaris}.
Similarly to starlings (and numerous other biological species), the proposed swarming system relies on sensing organs that look on two sides (cameras in our case), observing close-proximity neighbors only and responding to these sensory inputs by a local behavior which together forms a swarm intelligence that reaches beyond the abilities of a single particle.

The \ac{UVDAR} method tackles the problem of mutual perception of swarm particles by localizing the bearing and the relative \threeD position of their artificial \ac{UV} light emission in time, using passive \ac{UV}-sensitive cameras.
The method is deployable in indoor and outdoor environments with no need for mutual communication or for a heavy-weight sensory setup.
In addition, it is real-time, low-cost, scalable, and easy to plug into existing swarm systems.
To verify the feasibility of the \ac{UVDAR} technique in an aerial communication-less swarm system, we employed \ac{UVDAR} to generate a decentralized bio-inspired swarming behavior employing local information about neighboring agents and close-proximity obstacles in real-world conditions.
As verified in real-world experiments, the proposed system for relative localization is accurate, robust, and reliable for use in decentralized local-information based swarming models.


\begin{figure*}
  \centering
  \begin{minipage}[t]{.48\textwidth}
    \includegraphics[width=1.0\columnwidth]{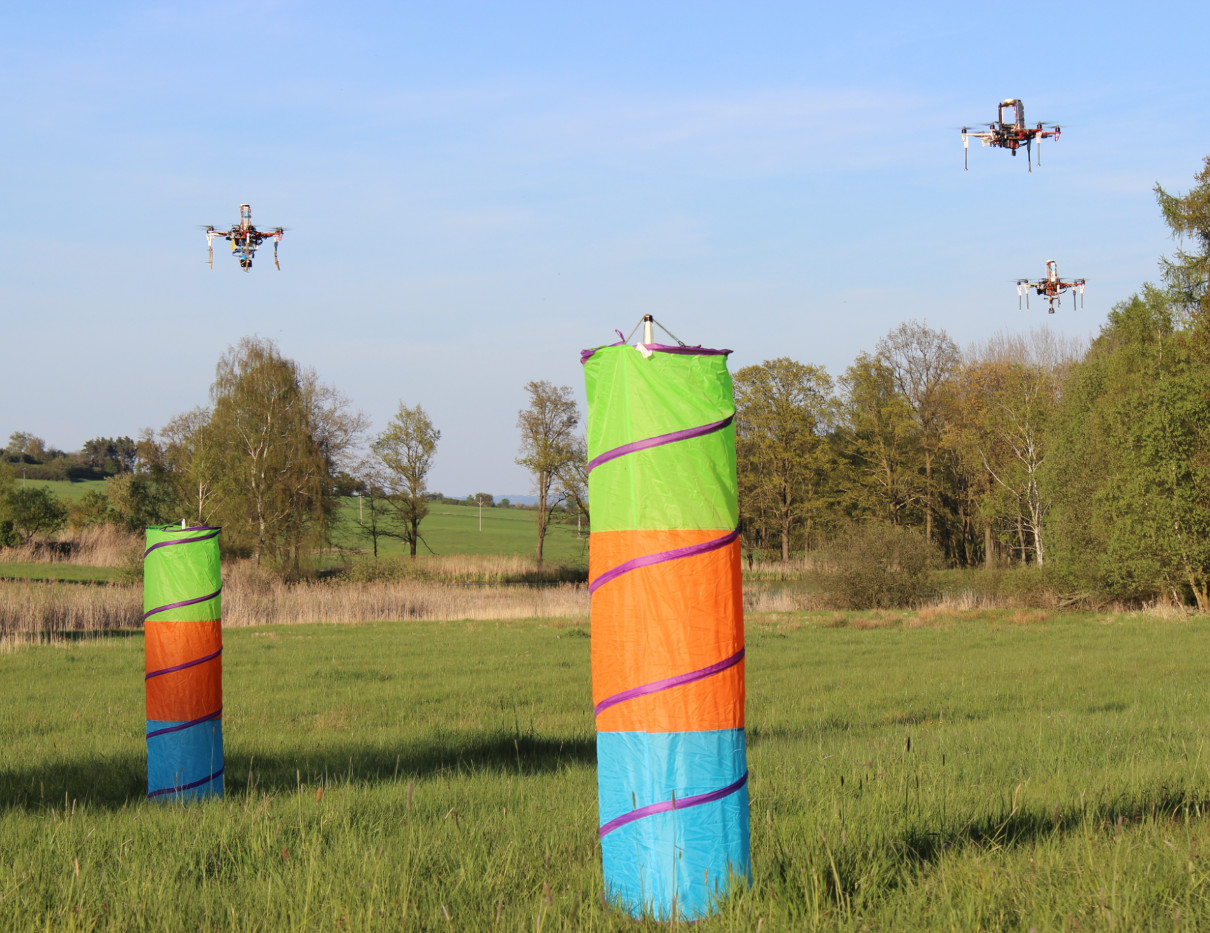}
    \caption{
      A compact aerial swarm of 3 {\acp{UAV}} in a controlled outdoor environment filled with artificial obstacles, as viewed by an outside observer.
      The decentralized approach, described in detail in \sref{sec:swarming}, applies a set of local rules contributing to safe navigation and self-organization of the swarm structure among obstacles. 
      The \acp{UAV} are homogeneous units with solely local sensing.
    }
    \label{fig:motivation_title_a}
  \end{minipage}\hspace{1em}
  \begin{minipage}[t]{.48\textwidth}
    \includegraphics[width=1.0\columnwidth]{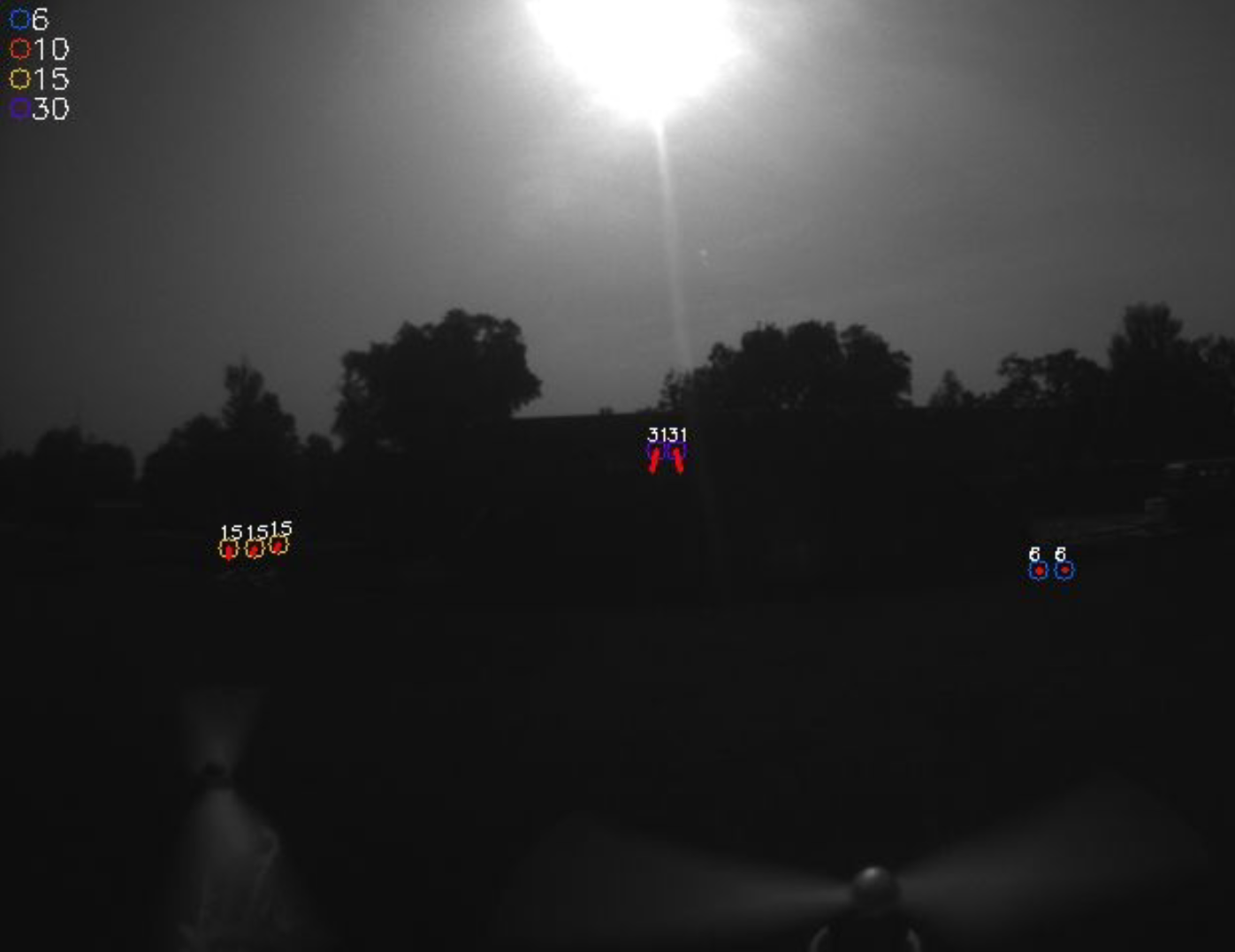}
    \caption{
      Onboard detection of 3 {\acp{UAV}} in the {\ac{UV}} spectrum using {\ac{UVDAR}} in a member of the aerial swarm.
    The method directly localizes the neighbors within a swarm in indoor and outdoor environments.
  Here, the method detects neighbors in an outdoor environment affected by a powerful source of ambient {\ac{UV}} radiation.
    The processing is possible due to periodic blinking of the members with a specific frequency, here with \SIlist{6;15;30}{\hertz}.
  }
    \label{fig:motivation_title_b}
  \end{minipage}
\end{figure*}



\subsection{Related Work}
\label{sec:sota}

\subsubsection{Relative Localization}
\hfill\vspace*{0.5em}\break
In most recent work concerning swarms and formation flight~\cite{swarm_overview}, the proposed algorithms have only been validated either in simulation or in laboratory-like conditions with the presence of absolute localization.
This was merely converted to relative measurements virtually, using systems such as \ac{RTK}-\ac{GNSS} or \ac{MOCAP}.
It is well known that \ac{MOCAP} is impractical for real-world deployment of mobile vehicles (either outdoors or indoors), as it requires the installation of an expensive infrastructure.
These absolute localization sources can provide the full pose of tracked objects, which oversimplifies the whole task \hl{with respect to} the reality of practical deployment.
Even if only partial information derived from absolute measurements is passed to the \acp{UAV} (e.g., distance or bearing), the continuous stream of such information is produced without realistic errors, which is unrepresentative of real-world conditions.

%

\hyphenation{in-fra-struct-ure}
Some more practical approaches consider infrastructure\-/less sensing such as ranging based on a radio signal~\cite{uwb}.
This only allows for distance-based following, without any orientation information, and requires a specific motion for sufficient state observability.
Another approach \cite{bt_directcommunication}, for the \twoD case, wirelessly  communicates the intentions of the leader.
This proves to be feasible since there are fewer degrees of freedom and there is less drift than in a general \threeD case.
These two approaches rely on radio transmission, which is subject to the effects of network congestion and interference.
For this reason, we consider vision-based approaches more suitable for multi-robot groups, especially in uncontrolled outdoor environments.


This approach has previously been explored by the authors' research group, relying on true outdoor relative localization, see~\cite{gpsdenied}.
The source of the relative localization was an onboard vision-based system using passive circular markers, as described in~\cite{whycon}.
There were, however, drawbacks: high sensitivity to the external lighting conditions and to partial occlusion, and substantial size for an acceptable detection range.

The use of active \ac{IR} markers has also been explored (see~\cite{ir_us_indoor,ir_indoor,scaramuzza_blinkers}) for the ability to suppress backgrounds using optical filtering.
These methods are however suitable solely for indoor, laboratory-like conditions, since solar radiation excessively pollutes the {\ac{IR}} spectrum, and subsequently the signal tends to deteriorate.
In~\cite{scaramuzza_blinkers}, the authors employed \ac{IR} markers with blinking frequency in the kilohertz range, which required event-based cameras to detect micro-scale changes.
These cameras are capable of detecting micro-scale changes.
However, they typically do not provide sufficiently high field of view and resolution, and they are not suitable for scalable swarms due to their size and cost.
The \ac{IR} spectrum has also been utilized in a passive manner~\cite{lf_ir_2donly}, but this approach, though simple, is even less robust to the outdoor conditions and distances applicable to \acp{UAV}.

It is also feasible to visually detect and localize unmarked \acp{UAV} using \ac{ML} methods such as \acp{CNN}.
However, these approaches require meticulously annotated datasets with a specific \ac{UAV} and with an environment similar to the intended operational space~\cite{cnn,8593405}.
The computational complexity and the dependency on satisfactory lighting conditions of such \ac{ML} systems precludes their deployment onboard lightweight \acp{UAV} suitable for swarming.
This motivated the development of the \ac{UVDAR} system, which is more robust to real-world conditions, because it reduces the computational load by optically filtering out visual information that is not of interest.
In contrast to~\cite{cnn,8593405}, \ac{UVDAR} also provides target identities.
The whole sensor is small, lightweight, and does not depend on the external lighting conditions.

\subsubsection{System Architecture}
\hfill\vspace*{0.5em}\break
To date, deployments of real-world aerial teams have not used any of the methodologies of direct localization described here in order to deal with the mesh-communication between the team members or with the communication link with a centralization element.
The record in terms of the number of \acp{UAV} cooperating at the same time is currently held by Intel\textsuperscript{\textregistered}~\cite{intel} with its fleet of Shooting Star quad-rotors.
Intel's centralized solution performs spectacular artistic light shows.
However in Intel's arrangement, each team member follows a pre-programmed trajectory, relying on \ac{GNSS} and a communication link with a ground station.
A similar methodology is employed in~\cite{Vir_gh_2014,DBLP:journals/corr/VasarhelyiVSTSNV14, gabor2018}, where the authors deployed swarms of \acp{UAV} in order to verify bio-inspired flocking behaviors in known confined environments.
In comparison with~\cite{intel}, their methods are decentralized; however, the \acp{UAV} still communicate their global states obtained by \ac{GNSS} within a radio-frequency mesh network.
This is not a realistic assumption in most application scenarios.

Recent successful real-world deployments are summarized in \tref{tab:sota_table}.
Observe that some kind of communication (either ground station to unit or unit-to-unit) is employed in most of the related work.
The dependency on a communication network lowers the upper limit for swarm scalability, due to the bandwidth limitations, and significantly reduces the fault tolerance of the entire system.
The \ac{UVDAR} relative visual perception system, described in detail in \sref{sec:uvdar}, is designed to remove this dependency.
Its use may allow working swarm systems to mimic the local behavioral mechanisms found in biological systems, ranging from general flocking to leader-follower scenarios.

\begin{table*}[t]
  \centering
  \captionsetup{width=0.9\textwidth}
  \begin{tabular}{llllll}
    \toprule
    Work & Decentralized & Communication & Relative localization\\\midrule
    \emph{Intel\textsuperscript{\textregistered}} \cite{intel} & No & Yes$^*$ & Shared global position (WiFi)\\
    \emph{EHang, Inc.} \cite{ehang} & No & Yes$^*$ & Shared global position (WiFi)\\
    \emph{Hauert et. al} \cite{6095129} & Yes & Yes & Shared global position (WiFi)\\
    \emph{B\"urkle et. al} \cite{Burkle2011} & Yes & Yes & Shared global position (WiFi)\\
    \emph{Kushleyev et. al} \cite{Kushleyev2013} & No & Yes$^*$ & Shared global position (ZigBee)\\
    \emph{V{\'{a}}s{\'{a}}rhelyi et. al} \cite{Vir_gh_2014,DBLP:journals/corr/VasarhelyiVSTSNV14, gabor2018} & Yes & Yes & Shared global position (XBee)\\
    \emph{Weinstein et. al} \cite{8276634} & No & Yes$^*$ & Shared global position (WiFi)
    \\\midrule
    \emph{Stirling et. al} \cite{stirling} & Yes & Yes & \ac{IIR} ranging\\
    \emph{Nguyen et. al} \cite{uwb} & N/A & Yes & \ac{UWB}\\
    \emph{Nägeli et. al} \cite{6942701} & Yes & Yes & Visual markers\\
    This work & Yes & \textbf{No} & \ac{UVDAR}\\
    \bottomrule
  \end{tabular}
  \caption{A brief comparison of aerial swarm systems with successful recent deployments outside of laboratory-like conditions. Methods marked with $(^*)$ employ communication with a centralized ground station.}
  \label{tab:sota_table}
\end{table*}

\subsubsection{Swarm Stabilization}
\label{sec:sota_swarm_stabilization}
\hfill\vspace*{0.5em}\break
To enable short-term stabilization of an autonomous \ac{UAV}, an onboard \ac{IMU} directly measures its linear acceleration, the attitude and the angular rate, using a combination of accelerometers, gyroscopes, and magnetometers.
To obtain long-term stabilization of an \ac{UAV}, however, it is not sufficient to use only the onboard \ac{IMU}, due to the inevitable measurement noises and drifts.
It is common practice to provide an additional estimate of the state vector variables (typically position or velocity), which is fused together with all the inertial measurements.
  The most common approach is to estimate the global position using a \ac{GNSS}.
  However, \ac{GNSS} signal availability is limited strictly to outdoor environments, and the accuracy of \ac{GNSS} is affected by an error of up to \SI{5}{\metre}~\cite{Diggelen2015TheWF}.
  Although the accuracy can be improved to \SI{2}{\centi\metre} with the use of \ac{RTK}-\ac{GNSS}, this makes aerial swarms deployable solely in controlled environments and is in contradiction with the bio-mimicking premise, since precise global localization is uncommon in biological systems.
  Other common methods of state estimation are local, and they typically employ onboard laser- or vision-based sensors to produce local estimates of the state variables.
  Vision-based methods may compute the optical flow to estimate the velocity of the camera \hl{relative to} the projected image plane~\cite{7073497}, or may apply algorithms of \ac{SLAM} to visual data~\cite{doi:10.1002/rob.21506}.
  Laser-based sensors are mostly used to estimate the relative motion between two frames of generated point-cloud data~\cite{KohlbrecherMeyerStrykKlingaufFlexibleSlamSystem2011}.

  There are structurally two approaches for stabilizing a swarm in a decentralized manner.
  The first group of methods distributes the state estimates determined for individual self-stabilization throughout the swarm (see \tref{tab:sota_table}). 
  In addition to restricting the communication infrastructure, this methodology has a major dependency between the swarm density and the accuracy of the global localization (e.g., \ac{GNSS}).
  In addition, it requires knowledge of individual transformations amid the coordination frames for distributed local state estimation methods.
  The second group of methods does not adopt a communication network to distribute the state estimates, but rather estimates the states directly from the relative onboard observations.
  This approach makes the swarm independent from the infrastructure, but it makes direct detection, estimation, and decision making with limited information more challenging.
  As further shown in \sref{sec:system_architecture}, the developed framework is part of the second group, perceiving the local neighborhood with visual organs and deploying a swarm of \acp{UAV} in fully-decentralized manner.




  \subsubsection{\hl{Swarming without Communication}}
  \label{sec:sota_communication_free_swarms}
  \hfill\vspace*{0.5em}\break
  \hl{%
    Decentralized swarming models accounting for complete or partial absence of communication were explored exclusively for 2D systems in the past (this is also implied in }\tref{tab:sota_table}\hl{).
  The majority of the state-of-the-art works within this field are biologically-inspired and emphasize self-organizing behavior of large-scale swarms of simple units with highly limited sensory capabilities.
  Highlighted is the Beeclust~}\cite{beeclust}\hl{ approach, which uses probabilistic finite state machines and a primitive motion model to mimic the collective behavior of honeybees.
  The Beeclust can be applied to complex tasks where information exchange among units is not required, such as in underwater exploration using a swarm of underwater robots~}\cite{BODI2015819}\hl{.
  A different method}~\cite{8613876}\hl{ analyzes the aggregation of agents towards a common spatial goal while avoiding inter-agent collisions.
  The authors of}~\cite{8613876}\hl{ show that their method with limited sensing properties of the agents performs similarly to methods employing complete pose information. 
  All of these decentralized algorithms require some form of mutual relative localization (even limited to binary detections), making them suitable for the use of }\ac{UVDAR}\hl{ localization.
  Overall review of the 2D approaches is systematically described in}~\cite{swarm_robots_review}\hl{, which further highlights the lack of research focus in the field of aerial swarming in 3D space.%
  }



\subsection{Contributions}

\hl{%
This article addresses problems of the deployment of real-world aerial swarms with no allowed communication or position sharing.
This potential problem is overcome with the use of the novel vision-based }\ac{UVDAR}\hl{ system for direct mutual perception of team members.%
}
The stability of the \ac{UVDAR} system for use in aerial swarming is the outcome of thorough real-world experimental verification in an outdoor environment with and without obstacles.
The main features of this article are as follows:
\begin{enumerate}[(i)]
  \item It provides an enabling technology for swarm research, often bio-inspired, by introducing a system that achieves fundamental swarm properties, as defined in~\cite{Trianni2008EvolutionarySR}.
  \item It introduces the \ac{UVDAR} system as an off-the-shelf tool for relative localization and identification of teammates suited for mutual perception of agents in robotic systems, such as aerial swarms.
  \item It introduces a decentralized bio-inspired swarming approach suited for obstacle-filled real-world environments, which requires only local relative information and no mutual communication.
  \item It verifies the feasibility and analyses the usability of aerial flocking relying on direct localization, which is the most frequent mechanism in biological systems.
  \item It is based on several real-world deployments of aerial swarms.
  \item It presents, to the best of our knowledge, the first autonomous deployments of aerial swarms with no centralized element and no mutual communication.
  \item It discloses the entire system as open source at \url{https://github.com/ctu-mrs}.
\end{enumerate}


\section{Motivation}




The lack of a communication-independent approach has put a constraint on much of the work done until now in the field of deploying teams of unmanned vehicles in challenging environments.
Our work here is motivated by the need for a communication-independent approach, and presents solutions that we have developed.
The insights into the development of the real-world deployments presented here tackle the motivations and constraints of the vast majority of related work restrained by the heretofore lack of communication-independent approaches.

Focusing on dense swarms of \acp{UAV} with short mutual distances, most of the swarming approaches reported in the literature have not been tested in real-world conditions.
Theoretical derivations, software simulations, and occasional experiments in laboratory conditions have formed the target for most of the related literature, as analyzed in~\cite{swarm_overview} and \cite{OH201783}.
However, this research milestone is far away from a meaningful real-world verification needed for an applicability of aerial swarms.
Real world interference cannot be neglected, as the integration of a swarming intelligence onto a multi-robot system yields constraints that need to be characterized directly in models of swarming behavior.

Instigated by biologically-inspired swarming models \cite{Smith_2019, OH201783} capable of achieving complex tasks (e.g., navigation, cohesion, food scouting, nest guarding, and predator avoidance) with a team of simple units, our aim was to imitate these models with the use of local information, as is widely observed in nature.
To allow the deployment of an infrastructure-independent (communication, environment) model, we had identified the most crucial factor impeding this type of deployment of a decentralized architecture -- the mutual relative localization between team members, which is also the most crucial information for animals in flocks in nature.
This motivated the development of the \ac{UVDAR} system (see \sref{sec:uvdar}), designed as a light-weight off-the-shelf plugin providing the local localization of neighboring swarm particles.
The usability of \ac{UVDAR} in dense swarms is analyzed in detail in \sref{sec:experimental_analysis}.
\section{UVDAR}
\label{sec:uvdar}

\begin{figure*}
  \begin{center}
   \captionsetup{width=1.0\textwidth}
    \input{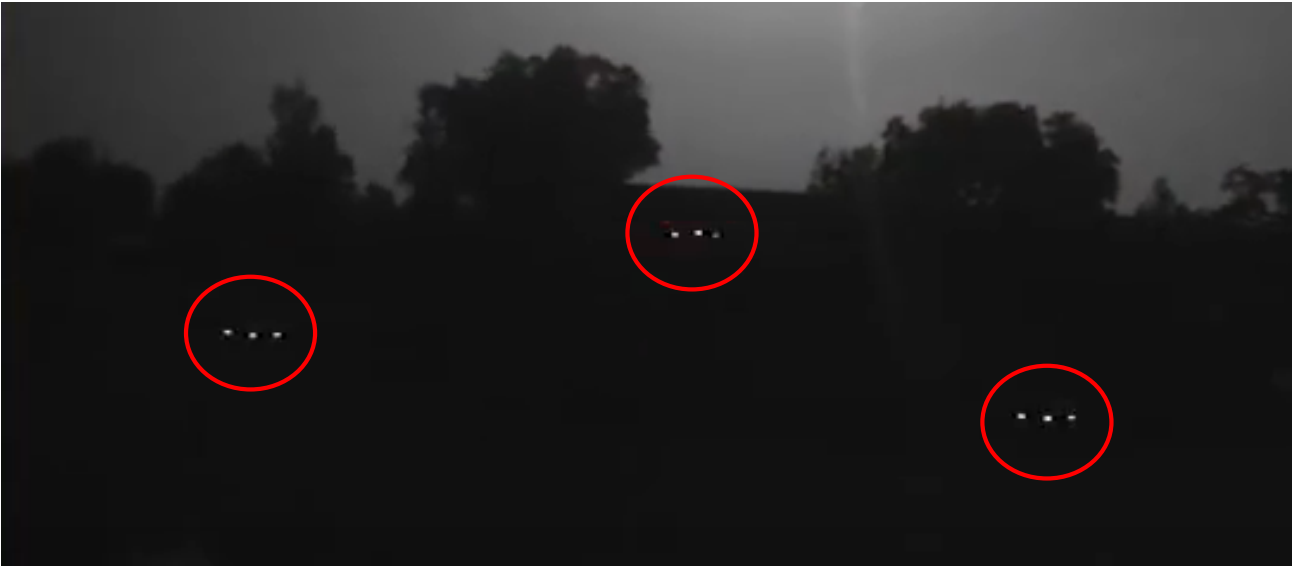}
    \caption{
      An example of the unprocessed view from the \ac{UV}-sensitive camera as a part of \ac{UVDAR} in a member of an aerial swarm.
      Note the extreme contrast of the LED markers in comparison to the background.
      A combination of the specific blinking frequency of the LED markers and the high contrast makes them simple to extract from background for processing.%
    }
    \label{fig:uvdar_raw}%
  \end{center}
\end{figure*}

Inspired by our extensive prior experimental experience with vision-based relative localization of \acp{UAV} (see \cite{whycon,visualstabilization}), we developed a novel relative localization sensor that tackles various limitations of previous solutions, namely the unpredictability of outdoor lighting and limits on the size and weight of onboard equipment.
The sensor, named \emph{\ac{UVDAR}}, is a \ac{UV} vision-based system comprising a \ac{UV}-sensitive camera and active \ac{UV} LED markers.
These lightweight, unobtrusive markers, attached to extreme points of a target \ac{UAV}, are seen as unique bright points in the \ac{UV} camera image (see \autoref{fig:uvdar_raw}).
This allows computationally simple detection \cite{uvdd2} and yields directly the relative bearing information of each marker from the perspective of the camera.
The fish-eye lenses that are used with the \ac{UV} camera provide a \SI{180}{\degree} horizontal overview of the surroundings.
Known camera calibration, together with the geometrical layout of the markers on the target, allows us also to retrieve an estimate of the distance (see \cite{uvdd2, uvdar_ral} for details).

In order to provide specific markers that would be distinguishable from others, and also to provide a further increase in robustness \hl{with respect to} outliers, we set the markers to blink with a specific sequence.
Using our specialized implementation of the \threeD time-position Hough transform (see \cite{uvdd2} for details), we can retrieve this signal for each observed marker, giving them identities.
In this project, we use these IDs to simplify the separation of multiple observed neighbor \acp{UAV}, but they can also be used to retrieve the relative orientation of the neighbors \cite{uvdar_ral}.
In addition to the swarming application described in this paper, {\ac{UVDAR}} may be used for e.g., a directed leader-follower flight~\cite{uvdar_ral}, where the use of the retrieved orientation is essential.
In addition, the neighbors' orientation estimate can be exploited for automatic generation of a dataset for training \ac{ML} vision for {\ac{UAV}} detection, as applied  in~\cite{uvdar_dataset_paper}, where {\ac{UVDAR}} was used for annotating color camera images.

\hl{%
  In swarms and in multi-{\ac{UAV}} systems in general, the blinking frequency of the onboard LEDs can be configured to encode information for optical data transmission between swarm units, in addition to using LED blinking directly for relative localization.%
  }
An example of such an application is in exploration, where a scouting unit can indicate the presence and the relative position of a discovered target to other units by combining various blinking signals and the unit's own orientation.
A further use is in cooperative voting in a group, where each unit expresses the current selection with blinking signals, and adjusts its vote on the basis of observing the selections of others.

In this paper, we go beyond our preliminary works with {\ac{UVDAR}}~\cite{uvdd2,uvdar_dataset_paper,uvdar_ral}, and also beyond other state-of-the-art literature, by incorporating direct mutual localization of {\acp{UAV}} \hl{into the position control feedback loop of a fully-decentralized swarming system without any kind of communication and external localization.
To the best of our knowledge, this paper presents the first real-world deployments of fully-decentralized bio-inspired swarms of {\acp{UAV}} using direct local localization for collective navigation in an uncontrolled environment.
}
This is what \ac{UVDAR} was intended for.

\subsection{\hl{Safety}}
\hl{The use of }\ac{UV}\hl{ radiation in the system has understandably raised some health concerns in the past.
We have verified the safety of this application by consulting the }\ac{ICNIRP}\hl{ "Guidelines on limits of exposure to ultraviolet radiation of wavelengths between 180 nm and 400 nm"}~\cite{UV2004guidelines}\hl{.
According to these guidelines, the exposure to }\ac{UV}\hl{ radiation (both to the eyes and to the skin) should not exceed }\SI{30}{\joule\per\meter\squared}\hl{ weighted by the relative spectral effectiveness (unitless wavelength-specific factor).
In our case of }\SI{395}{\nano\meter}\hl{ radiation, this factor equals to 0.000036, making the actual limit }\SI{8.3d5}{\joule\per\meter\squared}\hl{.
This means that our LEDs, producing }\SI{230}{\milli\watt}\hl{ of total radiated power}~\cite{led}\hl{ at the given driving current, can be safely viewed from the distance of }\SI{1}{\meter}\hl{ from the frontal direction (with the highest intensity in its Lambertian radiation pattern) for over }\SI{3000}{\hour}\hl{, making it effectively harmless.}


\subsection{Scalability}
\label{sec:uvdar_scalability}

In the context of a robotic swarm, scalability of the whole system is an important factor.
Using a communication network in large groups of robots limits the scalability by an upper bound defined by the total bandwidth, by the number of available channels, by the network architecture, or by the required data flow.
Employing a local perception method such as \ac{UVDAR}, the state of \textit{swarm particles} (team members, swarm units) is shared via direct observations, as is common in swarms in nature.
This system therefore does not need an explicit radio communication network.

As a vision-based method, \ac{UVDAR} suffers from natural restrictions, namely visual occlusions, camera resolution, and the detection, separation, and identification of image objects.
The upper scalability bound is determined by the ability to filter out the \ac{UV} markers belonging to a given swarm agent.
If the markers of all \acp{UAV} in the swarm are set to blink with the same frequency, individual agents have to be distinguished by separating their positions in the \ac{UV} image and in the constellations that they form.
In this case, we estimate that each agent should be capable of distinguishing up to 30 neighboring agents within the range of the {\ac{UVDAR}} system, bounded by the computational limitations.
This is however not the ideal mode of operation, as it becomes problematic when there are occlusions between agents, or when the agents are in close proximity in the observed image.

To tackle this challenge, we apply different blinking frequencies to different agents.
The \ac{UVDAR} system in its current configuration can accommodate up to 6 different frequencies of blinking that can be reliably distinguished from each other.
This allows us to mitigate the issue of overlapping agents - indeed, even agents that are directly behind each other can often be separated, if extreme markers of the further agent protrude into the image.
However, since the number of usable blinking frequencies is limited, we need to devise a method for spreading them evenly in the swarm, such that the likelihood of image separation of overlapping agents based on different frequencies between them is maximized for the whole swarm.
This has to be done in a decentralized manner, in order not to violate the swarming paradigm.

\hl{%
  One way to solve this for dense }\ac{UAV}\hl{ swarms is to have each agent dynamically re-assign its blinking frequency to differ as much as possible from the neighbors that it observes.
  This challenge definition can be likewise defined as the constraint satisfaction problem solved within a decentralized swarm of }\acp{UAV}\hl{ using direct observations only.
  The idea of this method is to maximize the local frequency diversity and additionally to allow all of the agents to initiate with the same ID (encoded by the blinking frequency of onboard markers).
  This opposes the current methodology of manually pre-setting the frequencies before deployment (see }\sref{sec:experimental_analysis}\hl{).
  The analysis and the theoretical limits on the convergence of such an approach towards a stable final state maximizing the scalability bound is still underway.}

  \hl{Another approach to increase the scalability bound, while carrying the identical ID on all the agents, lies in the design of }\ac{UVDAR}\hl{ itself.
  It is possible to introduce an additional omnidirectional }\ac{UV}\hl{ source on top of each agent. 
  This additional source is called a \textit{beacon} and it blinks with a specific frequency unique to the rest of the onboard markers on an agent.
  This allows for the separation of pixels in the image stream based on their image distance as well as their association with the singular beacon marker.
  The presence of at least two beacons in one region of the observer's image clearly implies a partial mutual occlusion.
  The use of beacons hence provides a limited ability to separate even agents in partial mutual occlusion relative to an observer if the beacons of both agents are visible.
}

The maximum range of detection should be taken into account for scalability in the geometrical sense.
With the current \ac{UVDAR} setup, detection is possible for targets up to \SI{15}{\meter} away from the sensor.
However, for improved reliability and robustness, a maximum range of \SI{10}{\meter} is recommended.
For determining the theoretical accuracy and range limitations, see~\cite{uvdd2}.
\hl{For a quantitative analysis on real-world accuracy, see }\sref{sec:direct_observation_accuracy}.
Filtering out distant targets, the limited detection range makes the method suitable for dense swarms, which place emphasis on a number of entities in a local neighborhood rather than on the swarm as a whole.
In biological systems, this perception characteristic allows for swarms of utmost magnitude, such as fish schools~\cite{Calovi_2014} with thousands of entities.




\section{Swarming Intelligence}
\label{sec:swarming}


In this article, we follow the swarm concept defined in \sref{sec:introduction}, in which the group is composed of swarm units with limited computational power and a short-term memory.
The concept is decentralized and uses autonomous self-organizing groups of homogeneous aerial vehicles operating in a \threeD space.

The proposed flocking approach works entirely with local information, with no requirement for any form of radio communication between the homogeneous swarm particles, and in an environment with convex obstacles.
The approach is inspired by biological systems, where global cooperative behavior can be found to emerge from elementary local interactions.
We will show that this phenomenon of cooperative behavior may yield collision-free stabilization in cluttered environments, self-organization of the swarm structure, and an ability to navigate in tasks suited for real {\acp{UAV}}.
The proposed swarming framework is founded on previously developed models~\cite{Reynolds:1987:FHS:37401.37406, 1605401}, which have been enhanced to suit the demands of real-world interference by extending them with concepts of obstacle avoidance, perception, and navigation.
\hl{The introduction of such extension concepts is highly important as the assumptions of dimensionless particles and an ideal world as in}~\cite{Reynolds:1987:FHS:37401.37406, 1605401}\hl{ do not apply in the real world.}
\hl{The main idea of the swarming behavior presented here} is to verify the feasibility, to perform an analysis, and to derive the properties of the \ac{UVDAR} system for use in swarm systems.
Bear in mind that \ac{UVDAR} is a general system and any swarming model~\cite{Vir_gh_2014,8613876, Zhu2019DistributedMF}, formation control approach~\cite{DBLP:journals/corr/abs-1904-03742}, or obstacle/predator avoidance method~\cite{Curiac_2015} utilizing local relative information can be employed to generate intelligent behavior when employing the \ac{UVDAR} system.

\subsection{Behavior Generation}
The behavioral model used throughout this article is defined in discrete time step~$k$ for a homogeneous swarm unit $i$ with an observation radius $R^{i}_{n}$~$\in$~$\mathbb{R}^{>0}$, an obstacle detection radius $R^{i}_{o}$~$\in$~$\mathbb{R}^{>0}$, a swarming velocity $\vect{v}_{[k]}^i$~$\in$~$\mathbb{R}^{3 \times 1}$, and a set of locally detected neighbors $\mset{N}^i_{[k]}$ within the observation radius $R^i_n$, as follows.
Bear in mind that all the relative observations in particle $i$ are given in the body frame of particle~$i$ at time step~$k$.%

The individual detected neighbor particles $j$~$\in$~$\mset{N}^i_{[k]}$ are represented by vectors of relative position $\vect{x}_{[k]}^{ij} \in \mathbb{R}^{3 \times 1}$ and relative velocity $\vect{v}_{[k]}^{ij} \in \mathbb{R}^{3 \times 1}$, $\forall j \in \{1,\,\dots,\,|\mset{N}^i_{[k]}|\}$, defined as
\begin{align}
  \vect{x}_{[k]}^{ij} &= \left[x_{[k]}^{ij}, y_{[k]}^{ij}, z_{[k]}^{ij}\right]^T,\label{eq:swarming_model_state_x}\\
  \vect{v}^{ij}_{[k]} &= \frac{1}{\Delta t^{ij}_{[k]}}\left(\vect{x}^{ij}_{[k]} - \vect{x}^{ij}_{[k-1]}\right) - \vect{v}_{[k-1]}^{i},\label{eq:swarming_model_state_v}
\end{align}
where $x_{[k]}^{ij}, y_{[k]}^{ij}, z_{[k]}^{ij}$ are Cartesian coordinates of a neighbor particle~$j$ represented in the body frame of agent $i$ in time step $k$, $\Delta t^{ij}_{[k]} = t^{ij}_{[k]} - t^{ij}_{[k-1]}$ is the time elapsed since the last direct detection of neighbor~$j$, and $\vect{v}^{i}_{[k=0]} = \vect{v}^{ij}_{[k=0]} = \vect{0}$.
The swarming model is then defined as a sum of elementary forces
\begin{equation}
  \vect{f}^{i}_{[k]}\left(\mset{N}^i_{[k]}, \mset{O}^i_{[k]}\right) =\;\vect{f}^{b,i}_{[k]}\left(\mset{N}^i_{[k]}\right) + \vect{f}^{n,i}_{[k]}\left(\mset{N}^i_{[k]}, \mset{O}^i_{[k]}\right),
  \label{eq:swarming_model}
\end{equation}
where $\vect{f}^{b,i}_{[k]}\left(\cdot\right) \in \mathbb{R}^{3 \times 1}$ embodies the baseline forces as an interpretation of the \textit{Boids model}~\cite{Reynolds:1987:FHS:37401.37406} flocking rules \textit{cohesion}, \textit{alignment}, and \textit{separation}, modified for real \acp{UAV} as
  \begin{equation}
    \small
    \vect{f}^{b,i}_{[k]}\left(\mset{N}^i_{[k]}\right) =  \frac{1}{\left|\mset{N}^i_{[k]}\right|}\sum_{j=1}^{\left|\mset{N}^i_{[k]}\right|}\left[\vect{x}^{ij}_{[k]} + \frac{\vect{v}^{ij}_{[k]}}{\lambda} - \kappa\left(\vect{x}^{ij}_{[k]}, R_n^i\right)\vect{x}^{ij}_{[k]}\right].
  \end{equation}
  The scalar $\lambda~$[\si{\hertz}] is the update rate of direct localization (camera rate) and the weighting function
  \begin{equation}
    \kappa(\vect{x}, r) = \max\left(0; \frac{\sqrt{\edist{\vect{x}}}}{\edist{\vect{x}}} - \frac{\sqrt{r}}{r}\right)
  \end{equation}
  represents a nonlinear weight coefficient scaling the repulsion behavior by the mutual distance between two neighbors.
  As the original model~\cite{Reynolds:1987:FHS:37401.37406} was designed for swarms of dimensionless particles, function $\kappa(\cdot)$ is particularly important for a swarm of real \acp{UAV}, in order to prevent mutual collisions while maintaining flexibility of the swarm as a whole.
  The force $\vect{f}_{[k]}^{n,i}(\mset{N}^i_{[k]}, \mset{O}^i_{[k]})$~$\in$~$\mathbb{R}^{3 \times 1}$~in~\eref{eq:swarming_model} is an extension to the simple model~\cite{Reynolds:1987:FHS:37401.37406} in the form of an additional navigation rule in an environment composed of $\mset{N}^i_{[k]}$ and a set of obstacles $\mset{O}^i_{[k]}$ detected within the detection radius $R^i_o$.

  The navigation rule can exploit any local multi-robot planning method~\cite{Elamvazhuthi_2019, 20.500.11850/83978, 5946136} in order to optimize the swarm motion parameters and to prevent a deadlock situation, or can include an obstacle avoidance mechanism and a navigation mechanism by introducing them as additional simplistic rules.
  To provide an example of the system performance, we introduce a simple attraction force $\vect{v}^{n,i}_{[k]} \in \mathbb{R}^{3 \times 1}$ towards a specified goal, together with a local reactive obstacle avoidance rule.
  To represent the obstacles, we introduce the concept of a \textit{virtual swarm particle}, which efficiently replaces a general geometric obstacle by a virtual entity.
This dimensionless particle is represented by a state comprised of a \hl{position and velocity relative to} particle $i$, similarly as defined in \eref{eq:swarming_model_state_x} and \eref{eq:swarming_model_state_v}.
  The methodology for finding the state of a virtual swarm particle is derived in the following section.
  The navigation rule is then derived as
  \begin{equation}
    \small
    \vect{f}^{n,i}_{[k]}\left(\mset{O}^i_{[k]}\right) = \frac{1}{\left|\mset{O}^i_{[k]}\right|}\sum_{v=1}^{\left|\mset{O}^i_{[k]}\right|}\left[\frac{\vect{v}^{iv}_{[k]}}{\lambda} - \kappa\left(\vect{x}^{iv}_{[k]}, R^i_o\right)\vect{x}^{iv}_{[k]}\right]
    + \frac{\vect{v}_{[k]}^{n,i}}{\lambda},
  \end{equation}
  where the vectors of the relative position $\vect{x}^{iv}_{[k]} \in \mathbb{R}^{3 \times 1}$ and the relative velocity $\vect{v}^{iv}_{[k]} \in \mathbb{R}^{3 \times 1}$ constitute the state of a $v$-th virtual swarm particle.

  \begin{figure*}[t]
    \centering
    \input{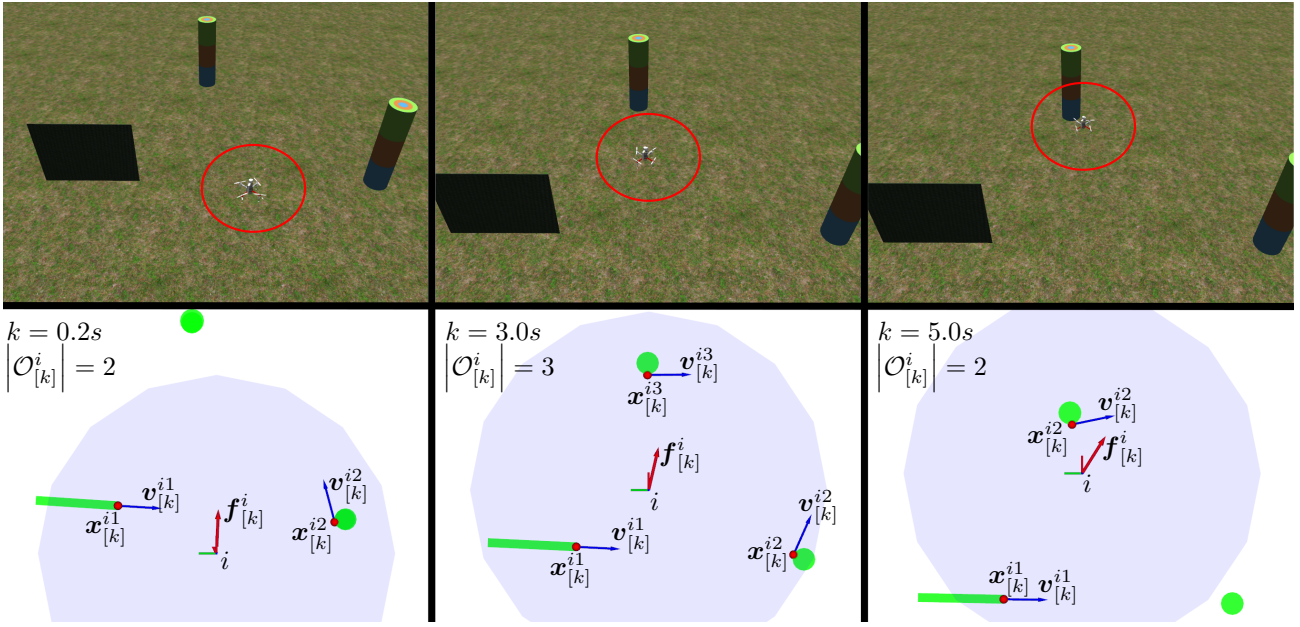}
    \caption{%
      An autonomous {\ac{UAV}} navigating among artificial obstacles according to the swarming model described in \sref{sec:swarming}.
      The {\ac{UAV}} flies in the Gazebo robotic simulator (upper row), while it continuously detects geometrical obstacles represented as circles and lines in the onboard \twoD laser-scanner data with a limited obstacle detection radius (gray circle).
        The states of virtual particles, consisting of \hl{position }$\vect{x}^{iv}_{[k]}$\hl{ (red dots) and velocity }$\vect{v}^{iv}_{[k]}$\hl{ (blue arrows) relative to} \ac{UAV} $i$, are visualized in the bottom image row.
        The steering force $\vect{f}^i_{[k]}$ (red arrow) of the swarming model represents the desired velocity.%
    }%
    \label{fig:obstacle_detection}
  \end{figure*}

  The swarming model defined in \eref{eq:swarming_model} represents the steering force of a particle $i$, which is used to compute the swarming velocity of particle $i$ as
  \begin{equation}
    \vect{v}_{[k]}^{i} = \gamma\left(\vect{f}_{[k]}^i\left(\mset{N}^i_{[k]}, \mset{O}^i_{[k]}\right)\right)\frac{\vect{f}_{[k]}^i\left(\mset{N}^i_{[k]}, \mset{O}^i_{[k]}\right)}{\left\lVert\vect{f}_{[k]}^i\left(\mset{N}^i_{[k]}, \mset{O}^i_{[k]}\right)\right\lVert_2},
  \end{equation}
  where
  \begin{equation}
    \gamma\left(\vect{f}\right)= \min\left\{v_m;\;\lambda\left\lVert\vect{f}\right\lVert_2\right\}
  \end{equation}
  bounds the magnitude of the velocity below the maximum allowed speed $v_{m}\,$[$\si{\metre\per\second}$].
  The swarming velocity is then used in real-world applications to compute the desired position setpoint as
  \begin{equation}
    \vect{r}_{[k]}^{d,i} = \frac{\vect{v}_{[k]}^{i}}{\lambda}
  \end{equation}
  represented in the body frame of \ac{UAV} $i$.


\subsection{Obstacle Detection}




  To achieve flocking in the targeted environment (e.g., a forest environment and an indoor environment), the obstacles in the local neighborhood are generalized into two geometrical classes (circles and lines), based on their cross-sections with the horizontal plane of a particle, as portrayed in \autoref{fig:obstacle_detection}.
This assumption allows us to model more complex settings (e.g., a forest or an office-like environment) on the grounds of these two geometrical classes, while it throttles down the perception and the computational complexity onboard a lightweight \ac{UAV}.
  Detection of these obstacles is assumed to be provided for a particle $i$ from any kind of an onboard sensor with an obstacle detection distance $R_o^i$.


  Having in time step $k$ a detected circular obstacle~$v$ with a radius~$r^v_{[k]} \in \mathbb{R}^{>0}$ and a center at~$\vect{c}^{iv}_{[k]} \in \mathbb{R}^{3 \times 1}$ referenced in the body frame of particle~$i$, the state of a $v$-th virtual swarm particle is derived as
  \begin{align}
    \vect{x}^{iv}_{[k]} &= \left(1 - \frac{
    r^v_{[k]}}{\edist{\vect{c}^{iv}_{[k]}}}\right)\vect{c}^{iv}_{[k]},\\
    \vect{v}^{iv}_{[k]} &= \frac{r^v_{[k]}}{\edist{\vect{c}^{iv}_{[k]}}}\left(\vect{I} - \vect{\mu}^{iv}_{[k]}\left(\vect{\mu}_{[k]}^{iv}\right)^T\right)\vect{v}^{i}_{[k]},
  \end{align}
  where $\edist{\cdot}$ is the L$^2$ norm, $\vect{I} \in \mathbb{R}^{3 \times 3}$ is an identity matrix, and $\vect{\mu}^{iv} = \frac{\vect{c}^{iv}}{\edist{\vect{c}^{iv}}}$.
  By analogy, the virtual swarm agent state can be derived for a linear obstacle defined by its normal vector $\vect{n}_{[k]}^{iv} \in \mathbb{R}^{3 \times 1}$ and a set of observed points $\mset{P}^{iv}_{[k]}$ as
  \begin{align}
    \vect{x}^{iv}_{[k]} &= (\vect{I} - \vect{P}_{[k]}^{iv})\,\hat{\vect{p}}^{iv}_{[k]}\\
    \vect{v}^{iv}_{[k]} &= \frac{1}{\edist{\hat{\vect{p}}^{iv}_{[k]}}}\vect{P}_{[k]}^{iv}\,\vect{v}^i_{[k]},
  \end{align}
  where 
  \begin{align}
    \vect{P}_{[k]}^{iv} &= \vect{I} - \vect{n}^{iv}_{[k]}\left(\vect{n}^{iv}_{[k]}\right)^{T},\\
    \hat{\vect{p}}^{iv}_{[k]} &= \arg\min_{\vect{p} \in \mset{P}^{iv}_{[k]}}\{\edist{\vect{p}}\}.
  \end{align}
  The state of a virtual swarm particle for both geometrical classes is visualized in \autoref{fig:obstacle_detection}, where an autonomous \ac{UAV} navigates among artificial obstacles within an environment of the Gazebo robotic simulator.


  

  \section{System Architecture}
  \label{sec:system_architecture}



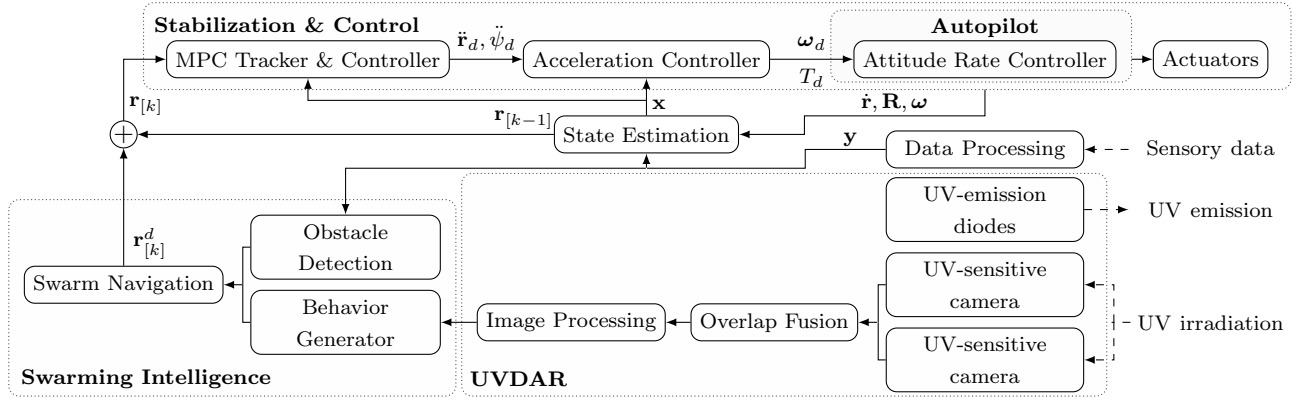
\begin{figure*}[t]
    \centering
    \usetikzlibrary{shapes.geometric,backgrounds,calc,arrows}
\usetikzlibrary{shadows}
\pgfdeclarelayer{background}
\pgfdeclarelayer{foreground}
\pgfsetlayers{background,main,foreground}

\tikzset{radiation/.style={{decorate,decoration={expanding waves,angle=90,segment length=4pt}}}}
\tikzstyle{block}=[draw, rounded corners, text centered, minimum height=1.4em, fill opacity=0.5, text opacity=1.0]
\tikzstyle{sum}=[draw, fill=gray!0, circle, inner sep=0pt, outer sep=0pt, node distance=1cm]

\begin{tikzpicture}[auto, |every node/.style={inner sep = .01em,outer sep=0.02em, text centered, minimum height = .5em}, >=latex]


  \node [block] (mpc) {\footnotesize MPC Tracker \& Controller};
  \node [block, right of=mpc, node distance=4.5cm] (so3) {\footnotesize Acceleration Controller};
  \node [block, right of=so3, fill=white, node distance=4.5cm] (attitude) {\footnotesize Attitude Rate Controller};
  \node [block, right of=attitude, node distance=3.0cm] (actuator) {\footnotesize Actuators};
  \node [block, below of=so3, node distance=1cm] (state) {\footnotesize State Estimation};
  \node [block, below of=attitude, node distance=1.2cm, text width=6.8em] (data_processing) {\footnotesize Data Processing};
  \node [block, below of=data_processing, node distance=0.8cm, text width=6.8em] (uvleds) {\footnotesize UV-emission diodes};
  \node [block, below of=uvleds, node distance=1.0cm, text width=6.8em] (camera1) {\footnotesize UV-sensitive camera};
  \node [block, below of=camera1, node distance=1.0cm, text width=6.8em] (camera2) {\footnotesize UV-sensitive camera};
  \node [block, left of=camera1, shift={(-1.8, -0.5)}] (uvdar_fusion) {\footnotesize Overlap Fusion};
  \node [block, left of=uvdar_fusion, node distance=2.7cm, shift={(0, -0.0)}] (uvdar_processing) {\footnotesize Image Processing};

  \node [rectangle, below of=actuator, node distance=1.2cm, text width=5.5em, text centered, shift = {(0.0, -0.0)}] (sensors) {\footnotesize Sensory data};
  \node [rectangle, below of=sensors, node distance=0.8cm, text width=5.5em, text centered, shift = {(0.0, -0.0)}] (emission) {\footnotesize UV emission};
  \node [rectangle, below of=emission, node distance=1.2cm, text width=6.5em, text centered, shift = {(0.0, -0.3)}] (radiation) {\footnotesize UV irradiation};

  \node [block, left of=uvdar_processing, node distance=3.0cm, text width=6.5em] (swarm_intelligence) {\footnotesize Behavior Generator};
  \node [block, left of=swarm_intelligence, shift = {(-1.95, 0.5)}] (swarm_navigation) {\footnotesize Swarm Navigation};
  \node [block, above of=swarm_intelligence, node distance=1.0cm, text width=6.5em] (obstacle_detection) {\footnotesize Obstacle Detection};
  
  \node [sum, above of=swarm_navigation, node distance=2.0cm] (plus) {+};



  \draw [->] (mpc.east) -- node[above, shift={(0, 0.00)}] {\footnotesize $\ddot{\mathbf{r}}_d, \ddot{\psi}_d$}(so3.west);
  \draw [->] (so3.east) -- node[above, shift={(-0.0, 0)}] {\footnotesize $\boldsymbol{\omega}_d$} node[below, shift={(-0.0, 0)}] {\footnotesize $T_d$}(attitude.west);
  \draw [->] (attitude.east)+(0.2,0) -- (actuator.west);

  \draw [->, dashed]  (sensors.west) -- (data_processing.east);
  \draw [->, dashed]  (uvleds.east) -- (emission.west);

  \draw [-, dashed]  (radiation.west)+(0.2,0)  -- +(-0.05,0);
  \draw [->, dashed] (radiation.west)+(-0.02,0) |- (camera1.east);
  \draw [->, dashed] (radiation.west)+(-0.02,0) |- (camera2.east);

  \draw [-]  (camera1.west)+(0.0,0)  -| +(-0.1,-0.5);
  \draw [-]  (camera2.west)+(0.0,0)  -| +(-0.1,0.5);
  \draw [->]  (uvdar_fusion.east)+(0.25,0)  -- +(0,0.0);
  \draw [->]  (uvdar_fusion.west)  -- (uvdar_processing.east);
  \draw [->]  (uvdar_processing.west)  -- (swarm_intelligence.east);

  \draw [->] (state.north) -- (so3.south);
  \draw [->] (state.north)+(0,0.20) node[right, shift={(-0.05, -0.05)}] {\footnotesize $\mathbf{x}$} -| (mpc.south);
  \draw [->] (state.west)+(0,0) node[above, shift={(-0.4, -0.05)}] {\footnotesize $\mathbf{r}_{[k-1]}$} -- (plus.east);
  \draw [->] (attitude.south)+(0,-0.15) |- node[above, shift={(-1.2, -0.08)}] {\footnotesize $\dot{\mathbf{r}}, \mathbf{R}, \boldsymbol{\omega}$} +(-2.4, -0.50) |- (state.east);

  \node[inner sep=0,minimum size=0,below of=state, node distance=0.45cm] (k) {}; 
  \draw [-] (data_processing.west) -| node[above, shift={(0.6, -0.08)}] {\footnotesize $\mathbf{y}$} +(-1.07, -0.0) |- (k);
  \draw [->] (k) -- (state.south);
  \draw [->] (k) -| (obstacle_detection.north);

  \draw [-]  (obstacle_detection.west)+(0.0,0) -| +(-0.1,-0.5);
  \draw [-]  (swarm_intelligence.west)+(0.0,0) -| +(-0.1,0.5);
  \draw [->]  (swarm_navigation.east)+(0.25,0)  -- +(0,0.0);
  \draw [->] (swarm_navigation.north) node[right, shift={(0.0, 0.30)}] {\footnotesize $\mathbf{r}^{d}_{[k]}$} -- (plus.south);
  \draw [->] (plus.north) node[right, shift={(-0.05, 0.2)}] {\footnotesize $\mathbf{r}_{[k]}$} |- (mpc.west);


  \begin{pgfonlayer}{background}
    \path (mpc.west |- mpc.north)+(-0.3,0.5) node (a) {};
    \path (actuator.south -| actuator.east)+(+0.3,-0.15) node (b) {};
    \path[fill=gray!1,rounded corners, draw=black!70, densely dotted]
    (a) rectangle (b);
  \end{pgfonlayer}
  \node [rectangle, above of=mpc, node distance=1.3em, shift={(-0.28,0.0)}] (text_control) {\footnotesize \textbf{Stabilization \& Control}};

  \begin{pgfonlayer}{background}
    \path (attitude.west |- attitude.north)+(-0.3,0.4) node (a) {};
    \path (attitude.south -| attitude.east)+(+0.2,-0.05) node (b) {};
    \path[fill=gray!3,rounded corners, draw=black!70, densely dotted]
    (a) rectangle (b);
  \end{pgfonlayer}
  \node [rectangle, above of=attitude, shift={(-0.0,0)}, node distance=1.2em] (autopilot) {\footnotesize \textbf{Autopilot}};


  \begin{pgfonlayer}{background}
    \path (uvdar_processing.west |- uvleds.north)+(-0.2,0.05) node (a) {};
    \path (camera2.south -| camera1.east)+(+0.3,-0.05) node (b) {};
    \path[fill=gray!1,rounded corners, draw=black!70, densely dotted]
    (a) rectangle (b);
  \end{pgfonlayer}
  \node [rectangle, below of=uvdar_processing, shift={(-0.7,0.25)}] (text_uvdar) {\footnotesize \textbf{UVDAR}};

  \begin{pgfonlayer}{background}
    \path (swarm_navigation.west |- obstacle_detection.north)+(-0.2,0.2) node (a) {};
    \path (camera2.south -| obstacle_detection.east)+(+0.2,-0.05) node (b) {};
    \path[fill=gray!1,rounded corners, draw=black!70, densely dotted]
    (a) rectangle (b);
  \end{pgfonlayer}
  \node [rectangle, below of=swarm_navigation, shift={(0.3,-0.25)}] (text_swarm) {\footnotesize \textbf{Swarming Intelligence}};


\end{tikzpicture}
    \caption[System Architecture]{
      The high-level system pipeline \hl{(the schematic is based on the system pipeline diagram published in}~\cite{8972370}\hl{)} of a single homogeneous \ac{UAV} swarm unit $i$ in time step $k$.
      The stabilization \& control pipeline~\cite{baca2020mrs} takes reference position setpoint $\mathbf{r}_{[k]}$ for the MPC in the MPC tracker, which outputs a command $\ddot{\mathbf{r}}_d,\;\ddot{\psi}_d$ ($\ddot{\psi}$ is the heading \hl{acceleration}) for the acceleration tracking $\mathrm{SO}(3)$ controller~\cite{5717652}.
      The acceleration controller produces the desired angular rate $\vect{\omega}_d$ and thrust reference $T_d$ for the embedded attitude rate controller.
      A state estimation pipeline outputs the current state estimate $\vect{x}$ based on the sensory data $\boldsymbol{y}$ and the onboard measurements of linear velocity $\dot{\vect{r}}$, angular rate $\boldsymbol{\omega}$, and attitude $\boldsymbol{R}$.
      Note that the time indices of the stabilization \& control and the state estimation pipelines are omitted in the diagram, since their timeline matches the rate of the inertial measurements (typically \SI{100}{\hertz}), which differs from the timeline of the detection cameras (\SIrange[range-phrase=--, range-units=single]{10}{20}{\hertz}).
      Local perception of neighboring units using the \ac{UVDAR} sensor is described in detail in \sref{sec:uvdar}, while the decentralized swarming approach is described thoroughly in \sref{sec:swarming}.
    }
    \label{fig:system_architecture}
  \end{figure*}

  In addition to the method for direct onboard localization presented in \sref{sec:uvdar} and the decentralized swarming approach presented in \sref{sec:swarming}, we will now present here system architecture of the entire \ac{UAV} system, supplemented by the concepts of \ac{UAV} stabilization, control, and state estimation.
  These concepts are based on our previous research (see~\cite{baca2020mrs,9024023,visualstabilization}) focused on cooperation among autonomous aerial vehicles.
  They have been adapted for swarming research described in this article.
  The control pipeline, suited for stabilizing and controlling \ac{UAV} swarms using linear \ac{MPC} and the non-linear $\mathrm{SO}(3)$ state feedback controller~\cite{5717652}, is depicted in the high-level scheme in \autoref{fig:system_architecture}.
  \hl{The stabilization and control pipeline is based entirely on}~\cite{baca2020mrs}.

  In addition, a decentralized collision avoidance system~\cite{baca2018mpc} is adapted in the proposed system for safe research on compact aerial swarms.
  A long prediction horizon of linear \ac{MPC} is used to detect collisions among trajectories of robots.
  The known collision trajectories are then altered prior their execution.
  This allows us to implement the collision avoidance system in a decentralized manner.
  Decentralized collision avoidance is necessary for safe verification of bio-inspired swarming models in the real world.
  Although the use of mutual communication for collision avoidance is in contradiction with the system architecture presented in this article, it can be used as a low-level safety supervisor with no direct dependency on the architecture of the tested swarming model.
  This may prevent inadmissible collisions when there is undesired demeanor of dense swarm members, and therefore protect the hardware during the initial phases of experimental swarm deployment.
  \hl{However, the use of collision avoidance is not mandatory and its use is appropriate only during the initial testing phase.}

  To stabilize \acp{UAV} using the system in \autoref{fig:system_architecture}, the individual \acp{UAV} estimate their state vector
  \begin{equation}
  \vect{x} = \left[\vect{r},\;\dot{\vect{r}},\;\ddot{\vect{r}},\;\vect{R},\;\boldsymbol{\omega}\right]^T,
  \label{eq:state_vector}
  \end{equation}
  where $\vect{R}$~$\in$~$\mathrm{SO}(3)$ is the attitude and $\vect{r}$~$=$~$\left[x_w, y_w, z_w\right]^T$ is the position in the world coordinate frame.
  The vector $\dot{\vect{r}}\in \mathbb{R}^{3 \times 1}$ is the linear velocity, $\ddot{\vect{r}}\in \mathbb{R}^{3 \times 1}$ is the linear acceleration, and $\boldsymbol{\omega}\in \mathbb{R}^{3 \times 1}$ is the angular rate \hl{with respect to} the \ac{UAV} body coordinate frame. 
  The PixHawk autopilot~\cite{10.1007/s10514-012-9281-4} is embedded to handle the low-level attitude rate and actuator control, and an \ac{IMU} is used to directly measure the linear acceleration~$\ddot{\vect{r}}$, the attitude~$\vect{R}$, and the angular rate~$\boldsymbol{\omega}$, using a combination of accelerometers, gyroscopes, and magnetometers.
  The embedded autopilot integrates the measurements of $\ddot{\vect{r}}$ to $\dot{\vect{r}}$ and employs the \ac{EKF} to produce optimal estimates of the specific state variables \hl{with respect to} the measurement noise.


  To self-localize an individual \ac{UAV}, its global position measured by \ac{GNSS} is fused together with the inertial measurements in order to stabilize the flight of this dynamically unstable system.
  However, the global state is not shared to other swarm agents throughout our final experimental analysis presented in \sref{sec:experimental_analysis}.
  Instead, the framework uses \ac{UVDAR} to directly observe the relative position and the relative velocity (see \eref{eq:swarming_model_state_x} and \eref{eq:swarming_model_state_v}) of particles in the local neighborhood, and it generates a navigation decision based on the set of simple rules described in \sref{sec:swarming}.
  Although the use of \ac{GNSS} for self-localization limits the system exclusively to outdoor environments, this dependency can be replaced by any local state estimation method \hl{with respect to} the desired application and environment -- e.g., the deployment of our decentralized system in a real-world forest, which was highlighted by the IEEE Spectrum~\footnote{\url{https://spectrum.ieee.org/automaton/robotics/drones/video-friday-dji-mavic-mini-palm-sized-foldable-drone}}.


  \subsection{\hl{Properties}}


  \hl{The combination of the system decentralization and the local perception of individual agents makes the system as a whole robust towards failures of individuals.
  In the swarming model (see }\sref{sec:swarming}\hl{), each agent decides on its actions in real time only from current observations or a short-past history of observations.
  This makes the system robust towards a single-point of failure, such as a failure of some centralized control element or the communication infrastructure.
  Unless the employed local perception method generates false negative detections, the swarming model (see }\sref{sec:swarming}\hl{) ensures no mutual collisions between the agents. The rate of false negative detections in UVDAR is minimal as there are no objects blinking at specific rates in the given near-visible UV spectrum.
  In case of a hardware failure of an aerial agent (e.g., the agent lands unexpectedly), the agent disappears from the visibility field of other units resulting in emergent self-organization of the collective configuration.

  As }\ac{UVDAR}\hl{ is a vision-based system, it naturally suffers from visual occlusions generating blind spots in overcrowded situations.
  As discussed in }\sref{sec:uvdar_scalability}\hl{, the number of visual occlusions in }\ac{UVDAR}\hl{ is mitigated with the use of different blinking frequencies of overlapping }\acp{UAV}\hl{. 
  As the neighborhood for perception is also locally limited in the swarming model (see }\sref{sec:swarming}\hl{), the distant blind spots are filtered out in principle.
  The remaining occluded agents are neglected.
  This is feasible in the employed model, as the information about the units' presence is propagated through direct observations of the motion of the middle agents (i.e., the agents causing the occlusions).
  Based on our empirical experience, this does not destabilize the swarm, but rather rearranges the agents to positions where the number of visual occlusions is reduced.


  The navigational features of the system as a whole are controlled in a decentralized manner.
  A decentralized navigation is possible with a swarming model capable of navigational decision making using only the perceived data onboard the units.
  This is the case of our swarming model (see }\sref{sec:swarming}\hl{), which employs a simple steering towards a pre-specified set of global positions, hence eliminating the need for navigation managed by a centralized controller.
  Although our later experiments (see }\sref{sec:experimental_analysis}\hl{) navigate each }\ac{UAV}\hl{ individually, the model may navigate only a single unit with the rest of the swarm naturally following the leader -- a behavior emerging from the cohesion and the alignment premises.}



  \begin{figure*}[t]
    \begin{center}
      \input{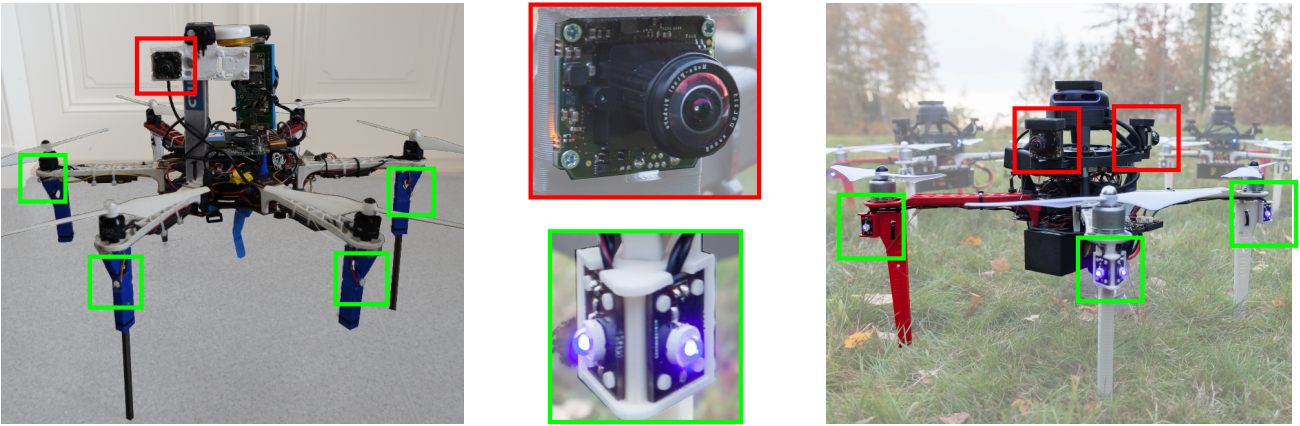}
      \caption{
          Two distinct multi-rotor (hexa- and quad-rotor) {\ac{UAV}} platforms, here equipped with {\ac{UV}}-sensitive cameras (red) and with active {\ac{UV}} markers (green), comprising the hardware components of the \ac{UVDAR} system for relative localization of neighboring {\acp{UAV}}.
          The diagonal dimension (without propellers) of the platforms are \SI{550}{\milli\meter} (left) and \SI{450}{\milli\meter} (right).
          The hexa-rotor platform was used throughout our experimental verification presented in \sref{sec:experimental_analysis}.
      }
      \label{fig:mav_platform}
    \end{center}
  \end{figure*}

  \subsection{Hardware Platform}
  

  The use of \ac{UVDAR} is not dependent on the dimensions or the configuration of a multi-rotor platform.
  The payload (onboard equipment) requirements of a single-\ac{UAV} unit employing \ac{UVDAR} are:
  an autopilot,
  a self-localization source (e.g., a \ac{GNSS} receiver),
  1-2 \ac{UV}-sensitive cameras, computational power to control the flight and to process the data (one camera at \SI{20}{\hertz} requires approximately a \SI{30}{\percent} single-thread load on Intel-Core i7 7567U, \SI{3.5}{\giga\hertz}), and
  a set of \ac{UV} LED markers placed at known extreme points of the \ac{UAV}.

  \hl{%
    To verify this statement, an axiomatic functionality validation of }\ac{UVDAR}\hl{ was performed on two independent multi-rotor platforms as shown in }\autoref{fig:mav_platform}\hl{.
  The general hardware configuration of }\acp{UAV}\hl{ exhibited in the figure consists of}
  \begin{itemize}
    \item \hl{the Pixhawk 4 autopilot,}
    \item \hl{onboard computer Intel NUC i7 7567U,}
    \item \hl{\textit{ProLight Opto PM2B-1LLE} near-}\ac{UV}\hl{ LEDs radiating at }\SIrange[range-phrase=--, range-units=single]{390}{410}{\nano\meter}\hl{ wavelength}~\cite{led}, 
    \item \hl{\textit{mvBlueFOX-MLC} cameras with}
      \begin{itemize}
        \item \hl{a \textit{MidOpt BP365} near-}{\ac{UV}}\hl{ band-pass filter and}
        \item \hl{\textit{Sunnex DSL215} fish-eye lenses,}
      \end{itemize}
    \item \hl{a }\ac{GNSS}\hl{ receiver (the hexa-rotor platform only), and}
    \item \hl{the Slamtec RPLiDAR-A3 laser scanner (the quad-rotor platform only).}
  \end{itemize}
  \hl{The weight of this hardware configuration is }\SI{370}{\gram}\hl{ (or }\SI{540}{\gram}\hl{ with the laser scanner required either for an obstacle detection or for a local localization replacing the }\ac{GNSS}\hl{ dependency).
  The onboard Intel NUC computer weighing }\SI{225}{\gram}\hl{ provides exaggerated processing power useful particularly in our case for general research purposes.
  For use in highly specialized applications, a feasible replacement of this payload with a microprocessor technology would allow for even further minimization of the aerial platform dimensions and cost expenses.

  Further miniaturization of }infrastructure\-/independent \acp{UAV}\hl{ is limited by current technology required for local self-localization.
  Vision-based algorithms employ lightweight cameras minimizing the weight; however, it comes at the cost of high processing power and thus increased weight of the processing unit.
  On the other hand, laser-based localization generally requires less processing power, but the sensors are heavier than cameras -- approximately }\SI{170}{\gram}\hl{ for planar scanners and }\SI{475}{\gram}\hl{ for 3D LiDARs.}



  \section{Experimental Analysis}
  \label{sec:experimental_analysis}

  \hl{%
    The primary aim of the experimental analysis is to verify the general functionality and to evaluate the performance of the entire framework exploiting direct localization rather than communication.
    The objectives of the experiments are focused primarily on determining the accuracy of the }\ac{UVDAR}\hl{ direct localization, and on the stabilization and spatial navigation of an aerial swarm in real-world environments with and without obstacles.}
  The entire experimental analysis is supported by multimedia materials available at \url{http://mrs.felk.cvut.cz/research/swarm-robotics}.


  \subsection{Swarming Model Analysis}
  \label{ssec:exp_swarming}

  To rule out the influence of \ac{UVDAR} in a position control feedback loop of an aerial swarm, the \textit{Boids}-based swarming intelligence (see \sref{sec:swarming}) is analyzed independently from the direct localization.
 For this purpose, the {\acp{UAV}} replace direct visual localization by sharing their global {\ac{GNSS}} positions in an ad-hoc network in order to determine the relative arrangement in the local neighborhood.
  This configuration was necessary in order to deploy {\acp{UAV}} without direct localization using {\ac{UVDAR}}, as discussed in \sref{sec:sota}.
 The analysis showcases the usability of the proposed fully-decentralized swarming framework both in simulations and in real-world scenarios, and in environments with and without obstacles.
  The global positions of the obstacles are apriori available to the \acp{UAV}.

 First, the collective dynamics of the swarming model are analyzed thoroughly in the Gazebo robotic simulator~\cite{1389727}, shown in \autoref{fig:obstacle_detection}, coupled with the \ac{ROS}~\cite{ros}.
 This simulation environment emulates real-world physics, and allows us to use identical low-level controllers and state estimation methods (see \sref{sec:system_architecture}) for the real {\acp{UAV}} and also for the simulated \acp{UAV}, without simplifying assumptions.
 This makes the configuration ideal for effortless deployment of theoretical bio-inspired swarming approaches onto a group of real-world robots.
 Simulation deployment of a swarm of homogeneous units in a 3D environment with obstacles (see \autoref{fig:sim_obstacle_avoidance}) verifies the qualitative performance of the reactive obstacle avoidance methodology presented in \sref{sec:swarming}.
 The emerging collective dynamics show the properties of the 3D shape flexibility during navigation through a narrow passage and in collision-free bypassing of static obstacles.
 The properties of safe navigation and high flexibility are also showcased during the simulation deployment of a compact swarm of 9 homogeneous units in a dense 3D forest-like environment, according to \autoref{fig:sim_9mavs_in_forest}.


  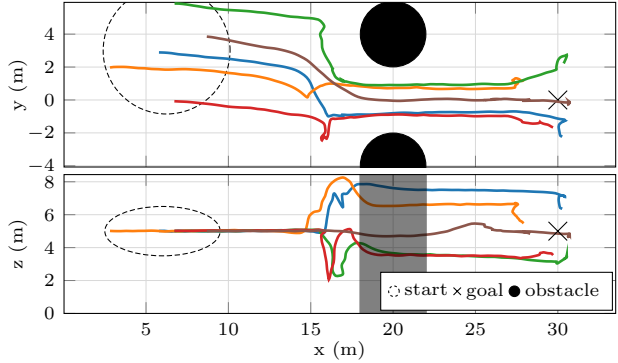
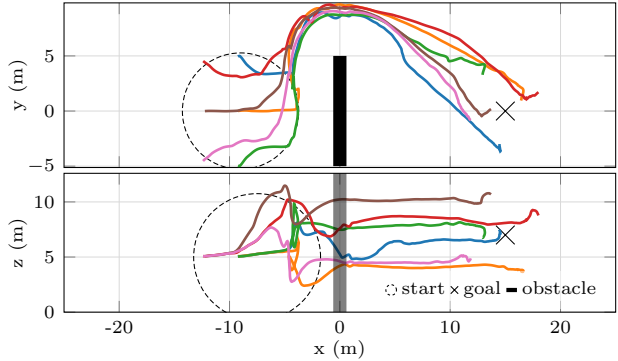
\begin{figure}[t]
    \centering
    \begingroup
    \captionsetup{width=0.95\columnwidth}
    \subfloat[Flexible and effective navigation of a decentralized swarm of 5 \acp{UAV} through a \SI{4}{\metre} wide narrow passage.]{




\begin{tikzpicture}[font=\scriptsize]
  \definecolor{color_uav1}{HTML}{1f77b4}
  \definecolor{color_uav2}{HTML}{ff7f0e}
  \definecolor{color_uav3}{HTML}{2ca02c}
  \definecolor{color_uav4}{HTML}{d62728}
  \definecolor{color_uav5}{HTML}{8c564b}
  \definecolor{color_uav6}{HTML}{e377c2}
  \definecolor{color_uav7}{HTML}{7f7f7f}
  \definecolor{color_uav8}{HTML}{bcbd22}
  \definecolor{color_uav9}{HTML}{17becf}

  \pgfplotstableread[col sep=space]{figure7a_uav1.txt}{\tableuavone}
  \pgfplotstableread[col sep=space]{figure7a_uav2.txt}{\tableuavtwo}
  \pgfplotstableread[col sep=space]{figure7a_uav3.txt}{\tableuavthree}
  \pgfplotstableread[col sep=space]{figure7a_uav4.txt}{\tableuavfour}
  \pgfplotstableread[col sep=space]{figure7a_uav5.txt}{\tableuavfive}

  \begin{axis}[ 
    name=top_view,
    width=0.52\textwidth,
    height=0.22\textwidth,
    grid=major,
    grid style={draw=gray!30,line width=.1pt},
    ylabel=y (m),
    y label style={at={(-0.05,0.5)}},
    axis equal,
    xtick={0, 5, 10, 15, 20, 25, 30},
    xticklabels={},
    enlarge y limits=false,
    axis line style={latex-latex},
    clip marker paths=true,
    ]
    
    \addplot+[only marks, mark=*, color=black, mark options={fill=black, scale=6.2}] table [y=y, x=x]{figure7a_obstacles.txt};

    \addplot [mark=o, color=black, mark options={scale=12.0}, dash pattern=on 2pt off 1pt on 2pt off 1pt] coordinates {(6.25, 3)};
    \addplot [mark=x, color=black, mark options={scale=2.5}] coordinates {(30, 0)};

    \addplot [smooth, color=color_uav1, line width=1pt] table [y=y, x=x]{\tableuavone};
    \addplot [smooth, color=color_uav2, line width=1pt] table [y=y, x=x]{\tableuavtwo};
    \addplot [smooth, color=color_uav3, line width=1pt] table [y=y, x=x]{\tableuavthree};
    \addplot [smooth, color=color_uav4, line width=1pt] table [y=y, x=x]{\tableuavfour};
    \addplot [smooth, color=color_uav5, line width=1pt] table [y=y, x=x]{\tableuavfive};

  \end{axis}

  \begin{axis}[ 
    name=side_view
    at=(top_view.below south west), anchor=above north west,
    width=0.52\textwidth,
    height=0.20\textwidth,
    grid=major,
    grid style={draw=gray!30,line width=.1pt},
    xlabel=x (m),
    ylabel=z (m),
    xtick={0, 5, 10, 15, 20, 25, 30},
    xticklabels={0, 5, 10, 15, 20, 25, 30},
    xlabel style={
    yshift=1.0ex},
    y label style={at={(-0.05,0.5)}},
    axis equal,
    ymin=0,
    enlarge y limits=false,
    legend columns=3,
    legend style={text=black, row sep=-0.4ex, at={(0.995, 0.3)}},
    legend entries={start, goal, obstacle},
    legend cell align={left},
    axis line style={latex-latex},
    ]
    \addlegendimage{only marks,mark=o, dash pattern=on 1pt off 1pt on 1pt off 1pt}
    \addlegendimage{only marks,mark=x}
    \addlegendimage{only marks,mark=*,mark options={solid, color=black}}

    \draw [black, dash pattern=on 2pt off 1pt on 2pt off 1pt] (6, 5) ellipse (3.5 and 1.5);
    \addplot [mark=x, color=black, mark options={scale=2.5}] coordinates {(30, 5)};

    \addplot[smooth, color=color_uav1, line width=1pt] table [y=z, x=x]{\tableuavone};
    \addplot[smooth, color=color_uav2, line width=1pt] table [y=z, x=x]{\tableuavtwo};
    \addplot[smooth, color=color_uav3, line width=1pt] table [y=z, x=x]{\tableuavthree};
    \addplot[smooth, color=color_uav4, line width=1pt] table [y=z, x=x]{\tableuavfour};
    \addplot[smooth, color=color_uav5, line width=1pt] table [y=z, x=x]{\tableuavfive};

    \draw[line width=25.5, black, opacity=0.5] (20, 0) -- (20, 20);
   
  \end{axis}
\end{tikzpicture}
    }\\\vspace*{-0.5em}
    \subfloat[Fast and efficient maneuvering of 6 \acp{UAV} emerging solely from local interactions during avoidance of a static obstacle.]{
      \begin{tikzpicture}[font=\scriptsize]
  \definecolor{color_uav1}{HTML}{1f77b4}
  \definecolor{color_uav2}{HTML}{ff7f0e}
  \definecolor{color_uav3}{HTML}{2ca02c}
  \definecolor{color_uav4}{HTML}{d62728}
  \definecolor{color_uav5}{HTML}{8c564b}
  \definecolor{color_uav6}{HTML}{e377c2}
  \definecolor{color_uav7}{HTML}{7f7f7f}
  \definecolor{color_uav8}{HTML}{bcbd22}
  \definecolor{color_uav9}{HTML}{17becf}

  \pgfplotstableread[col sep=space]{figure7b_uav1.txt}{\tableuavone}
  \pgfplotstableread[col sep=space]{figure7b_uav2.txt}{\tableuavtwo}
  \pgfplotstableread[col sep=space]{figure7b_uav3.txt}{\tableuavthree}
  \pgfplotstableread[col sep=space]{figure7b_uav4.txt}{\tableuavfour}
  \pgfplotstableread[col sep=space]{figure7b_uav5.txt}{\tableuavfive}
  \pgfplotstableread[col sep=space]{figure7b_uav6.txt}{\tableuavsix}

  \begin{axis}[ 
    name=top_view,
    width=0.52\textwidth,
    height=0.22\textwidth,
    grid=major,
    grid style={draw=gray!30,line width=.1pt},
    xtick={-20, -10, 0, 10, 20, 30},
    xticklabels={},
    ylabel=y (m),
    y label style={at={(-0.05,0.5)}},
    xmin=-25, xmax=25,
    axis equal,
    enlarge y limits=false,
    axis line style={latex-latex},
    clip marker paths=true,
    ]
    
    \draw[line width=5, black] (0, -5) -- (0, 5);

    \addplot [mark=o, color=black, mark options={scale=11.0}, dash pattern=on 2pt off 1pt on 2pt off 1pt] coordinates {(-9, 0)};
    \addplot [mark=x, color=black, mark options={scale=2.5}] coordinates {(15, 0)};

    \addplot [smooth, color=color_uav1, line width=1pt] table [y=y, x=x]{\tableuavone};
    \addplot [smooth, color=color_uav2, line width=1pt] table [y=y, x=x]{\tableuavtwo};
    \addplot [smooth, color=color_uav3, line width=1pt] table [y=y, x=x]{\tableuavthree};
    \addplot [smooth, color=color_uav4, line width=1pt] table [y=y, x=x]{\tableuavfour};
    \addplot [smooth, color=color_uav5, line width=1pt] table [y=y, x=x]{\tableuavfive};
    \addplot [smooth, color=color_uav6, line width=1pt] table [y=y, x=x]{\tableuavsix};
  \end{axis}

  \begin{axis}[ 
    name=side_view
    at=(top_view.below south west), anchor=above north west,
    width=0.52\textwidth,
    height=0.20\textwidth,
    grid=major,
    grid style={draw=gray!30,line width=.1pt},
    xlabel=x (m),
    ylabel=z (m),
    xlabel style={
    yshift=1.0ex},
    y label style={at={(-0.05,0.5)}},
    xtick={-20, -10, 0, 10, 20, 30},
    xticklabels={-20, -10, 0, 10, 20, 30},
    axis equal,
    ymin=0,
    enlarge y limits=false,
    xmin=-25, xmax=25,
    legend columns=3,
    legend style={text=black, row sep=-0.4ex, at={(0.99, 0.3)}, fill opacity=0.5, text opacity=1, draw=none},
    legend entries={start, goal, obstacle},
    legend cell align={left},
    axis line style={latex-latex},
    clip marker paths=true,
    yshift=-0.2em,
    ]
    \addlegendimage{only marks,mark=o, dash pattern=on 1pt off 1pt on 1pt off 1pt}
    \addlegendimage{only marks,mark=x}
    \addlegendimage{only marks,mark=-,mark options={solid, color=black, ultra thick}}

    \addplot [mark=o, color=black, mark options={scale=12.0}, dash pattern=on 2pt off 1pt on 2pt off 1pt] coordinates {(-7.5, 5)};
    \addplot [mark=x, color=black, mark options={scale=2.5}] coordinates {(15, 7)};

    \addplot[smooth, color=color_uav1, line width=1pt] table [y=z, x=x]{\tableuavone};
    \addplot[smooth, color=color_uav2, line width=1pt] table [y=z, x=x]{\tableuavtwo};
    \addplot[smooth, color=color_uav3, line width=1pt] table [y=z, x=x]{\tableuavthree};
    \addplot[smooth, color=color_uav4, line width=1pt] table [y=z, x=x]{\tableuavfour};
    \addplot[smooth, color=color_uav5, line width=1pt] table [y=z, x=x]{\tableuavfive};
    \addplot[smooth, color=color_uav6, line width=1pt] table [y=z, x=x]{\tableuavsix};

    \draw[line width=5, black, opacity=0.5] (0, 0) -- (0, 20);

  \end{axis}
\end{tikzpicture}
    }
    \endgroup
    \caption{
      A fully-decentralized swarm of homogeneous units in a simulated 3D environment with static obstacles.
      The swarming model yields enough flexibility for the compact team to deviate from its aggregated structure in order to pass safely through a narrow gap (a) or to avoid an obstacle in an efficient and fast manner (b).
    }
    \label{fig:sim_obstacle_avoidance}
  \end{figure}
  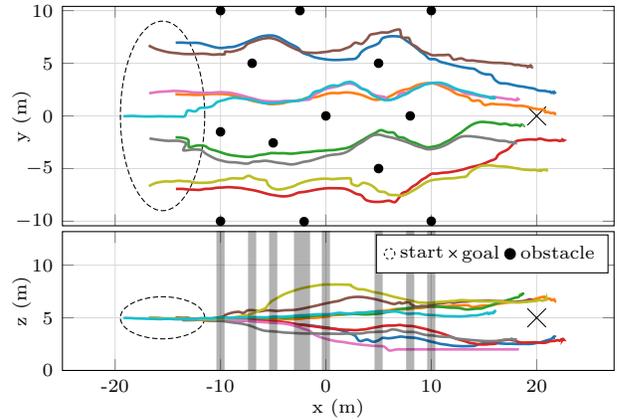
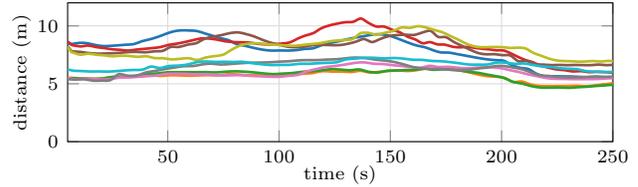
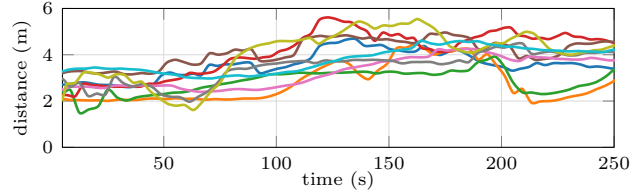
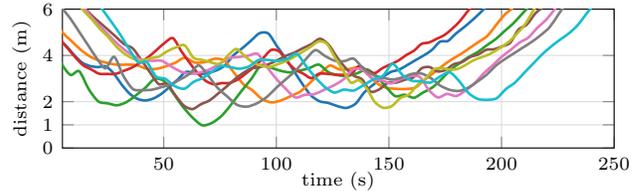
\begin{figure}[t]
    \centering
    \subfloat[Orthogonal views on the trajectories of the particles.]{
      \begin{tikzpicture}[font=\scriptsize]
  
  \definecolor{color_uav1}{HTML}{1f77b4}
  \definecolor{color_uav2}{HTML}{ff7f0e}
  \definecolor{color_uav3}{HTML}{2ca02c}
  \definecolor{color_uav4}{HTML}{d62728}
  \definecolor{color_uav5}{HTML}{8c564b}
  \definecolor{color_uav6}{HTML}{e377c2}
  \definecolor{color_uav7}{HTML}{7f7f7f}
  \definecolor{color_uav8}{HTML}{bcbd22}
  \definecolor{color_uav9}{HTML}{17becf}

  \pgfplotstableread[col sep=space]{figure8_uav1.txt}{\tableuavone}
  \pgfplotstableread[col sep=space]{figure8_uav2.txt}{\tableuavtwo}
  \pgfplotstableread[col sep=space]{figure8_uav3.txt}{\tableuavthree}
  \pgfplotstableread[col sep=space]{figure8_uav4.txt}{\tableuavfour}
  \pgfplotstableread[col sep=space]{figure8_uav5.txt}{\tableuavfive}
  \pgfplotstableread[col sep=space]{figure8_uav6.txt}{\tableuavsix}
  \pgfplotstableread[col sep=space]{figure8_uav7.txt}{\tableuavseven}
  \pgfplotstableread[col sep=space]{figure8_uav8.txt}{\tableuaveight}
  \pgfplotstableread[col sep=space]{figure8_uav9.txt}{\tableuavnine}

  \begin{axis}[ 
    name=top_view,
    width=0.52\textwidth,
    grid=major,
    grid style={draw=gray!30,line width=.1pt},
    xtick={-20, -10, 0, 10, 20, 30},
    xticklabels={},
    ylabel=y (m),
    y label style={at={(-0.04,0.5)}},
    axis equal image,
    xmin=-25,
    enlarge y limits=0.02,
    axis line style={latex-latex},
    clip marker paths=true,
    ]
    
    \addplot+[only marks, mark=*, color=black, mark options={fill=black, scale=0.8}] table [y=y, x=x]{figure8_obstacles.txt};

    \draw [black, dash pattern=on 2pt off 1pt on 2pt off 1pt] (-15.5, 0) ellipse (4 and 9);
    \addplot [mark=x, color=black, mark options={scale=2.5}] coordinates {(20, 0)};

    \addplot [smooth, color=color_uav1, line width=1pt] table [y=y, x=x]{\tableuavone};
    \addplot [smooth, color=color_uav2, line width=1pt] table [y=y, x=x]{\tableuavtwo};
    \addplot [smooth, color=color_uav3, line width=1pt] table [y=y, x=x]{\tableuavthree};
    \addplot [smooth, color=color_uav4, line width=1pt] table [y=y, x=x]{\tableuavfour};
    \addplot [smooth, color=color_uav5, line width=1pt] table [y=y, x=x]{\tableuavfive};
    \addplot [smooth, color=color_uav6, line width=1pt] table [y=y, x=x]{\tableuavsix};
    \addplot [smooth, color=color_uav7, line width=1pt] table [y=y, x=x]{\tableuavseven};
    \addplot [smooth, color=color_uav8, line width=1pt] table [y=y, x=x]{\tableuaveight};
    \addplot [smooth, color=color_uav9, line width=1pt] table [y=y, x=x]{\tableuavnine};

  \end{axis}

  \begin{axis}[ 
    name=side_view
    at=(top_view.below south west), anchor=above north west,
    width=0.52\textwidth,
    height=0.20\textwidth,
    grid=major,
    grid style={draw=gray!30,line width=.1pt},
    xtick={-20, -10, 0, 10, 20, 30},
    xticklabels={-20, -10, 0, 10, 20, 30},
    xlabel=x (m),
    ylabel=z (m),
    xlabel style={
    yshift=1.0ex},
    y label style={at={(-0.04,0.5)}},
    axis equal,
    xmin=-25,
    ymin=0,
    enlarge y limits=false,
    legend columns=3,
    legend style={text=black, row sep=-0.4ex, at={(0.99, 0.98)}},
    legend entries={start, goal, obstacle},
    legend cell align={left},
    axis line style={latex-latex},
    yshift=-0.2em,
    ]
    \addlegendimage{only marks,mark=o, dash pattern=on 1pt off 1pt on 1pt off 1pt}
    \addlegendimage{only marks,mark=x}
    \addlegendimage{only marks,mark=*,mark options={solid,color=black}}
   
    \draw [black, dash pattern=on 2pt off 1pt on 2pt off 1pt] (-15.5, 5) ellipse (4 and 2);
    \addplot [mark=x, color=black, mark options={scale=2.5}] coordinates {(20, 5)};

    \addplot [smooth, color=color_uav1, line width=1pt] table [y=z, x=x]{\tableuavone};
    \addplot [smooth, color=color_uav2, line width=1pt] table [y=z, x=x]{\tableuavtwo};
    \addplot [smooth, color=color_uav3, line width=1pt] table [y=z, x=x]{\tableuavthree};
    \addplot [smooth, color=color_uav4, line width=1pt] table [y=z, x=x]{\tableuavfour};
    \addplot [smooth, color=color_uav5, line width=1pt] table [y=z, x=x]{\tableuavfive};
    \addplot [smooth, color=color_uav6, line width=1pt] table [y=z, x=x]{\tableuavsix};
    \addplot [smooth, color=color_uav7, line width=1pt] table [y=z, x=x]{\tableuavseven};
    \addplot [smooth, color=color_uav8, line width=1pt] table [y=z, x=x]{\tableuaveight};
    \addplot [smooth, color=color_uav9, line width=1pt] table [y=z, x=x]{\tableuavnine};
 
    \draw[line width=3.1, black, opacity=0.3] (0, 0) -- (0, 20);
    \draw[line width=3.1, black, opacity=0.3] (-10, 0) -- (-10, 20);
    \draw[line width=6.1, black, opacity=0.3] (-2.2665, 0) -- (-2.2665, 20);
    \draw[line width=3.1, black, opacity=0.3] (10, 0) -- (10, 20);
    \draw[line width=3.1, black, opacity=0.3] (5, 0) -- (5, 20);
    \draw[line width=3.1, black, opacity=0.3] (-7, 0) -- (-7, 20);
    \draw[line width=3.1, black, opacity=0.3] (-5, 0) -- (-5, 20);
    \draw[line width=3.1, black, opacity=0.3] (8, 0) -- (8, 20);

  \end{axis}
\end{tikzpicture}
    }\\\vspace*{-0.5em}
    \subfloat[Average distance amid the UAVs.]{
      \begin{tikzpicture}[font=\scriptsize]
  \definecolor{color_uav1}{HTML}{1f77b4}
  \definecolor{color_uav2}{HTML}{ff7f0e}
  \definecolor{color_uav3}{HTML}{2ca02c}
  \definecolor{color_uav4}{HTML}{d62728}
  \definecolor{color_uav5}{HTML}{8c564b}
  \definecolor{color_uav6}{HTML}{e377c2}
  \definecolor{color_uav7}{HTML}{7f7f7f}
  \definecolor{color_uav8}{HTML}{bcbd22}
  \definecolor{color_uav9}{HTML}{17becf}

  \begin{axis}[ 
    width=0.515\textwidth,
    height=0.2\textwidth,
    grid=major,
    grid style={draw=gray!30,line width=.1pt},
    xlabel=time (s),
    ylabel=distance (m),
    xlabel style={
    yshift=1.4ex},
    y label style={at={(-0.05,0.5)}},
    xmin=5, xmax=250,
    ymin=0, ymax=12,
    xtick={2, 50, 100, 150, 200, 250},
    xticklabels={0, 50, 100, 150, 200, 250},
    axis line style={latex-latex},
    legend columns=1,
    legend style={text=black, at={(0.995, 0.12)}},
    legend image post style={black}
    ]
    


    \addplot [smooth, color=color_uav1, line width=1pt] table [y=avg_mav_dist, x=time]{figure8_uav1.txt};
    \addplot [smooth, color=color_uav2, line width=1pt] table [y=avg_mav_dist, x=time]{figure8_uav2.txt};
    \addplot [smooth, color=color_uav3, line width=1pt] table [y=avg_mav_dist, x=time]{figure8_uav3.txt};
    \addplot [smooth, color=color_uav4, line width=1pt] table [y=avg_mav_dist, x=time]{figure8_uav4.txt};
    \addplot [smooth, color=color_uav5, line width=1pt] table [y=avg_mav_dist, x=time]{figure8_uav5.txt};
    \addplot [smooth, color=color_uav6, line width=1pt] table [y=avg_mav_dist, x=time]{figure8_uav6.txt};
    \addplot [smooth, color=color_uav7, line width=1pt] table [y=avg_mav_dist, x=time]{figure8_uav7.txt};
    \addplot [smooth, color=color_uav8, line width=1pt] table [y=avg_mav_dist, x=time]{figure8_uav8.txt};
    \addplot [smooth, color=color_uav9, line width=1pt] table [y=avg_mav_dist, x=time]{figure8_uav9.txt};

  \end{axis}
\end{tikzpicture}
    }\\\vspace*{-1.0em}
    \subfloat[Minimal distance amid the UAVs.]{
      \begin{tikzpicture}[font=\scriptsize]
  \definecolor{color_uav1}{HTML}{1f77b4}
  \definecolor{color_uav2}{HTML}{ff7f0e}
  \definecolor{color_uav3}{HTML}{2ca02c}
  \definecolor{color_uav4}{HTML}{d62728}
  \definecolor{color_uav5}{HTML}{8c564b}
  \definecolor{color_uav6}{HTML}{e377c2}
  \definecolor{color_uav7}{HTML}{7f7f7f}
  \definecolor{color_uav8}{HTML}{bcbd22}
  \definecolor{color_uav9}{HTML}{17becf}

  \begin{axis}[ 
    width=0.52\textwidth,
    height=0.2\textwidth,
    grid=major,
    grid style={draw=gray!30,line width=.1pt},
    xlabel=time (s),
    ylabel=distance (m),
    xlabel style={
    yshift=1.4ex},
    y label style={at={(-0.04,0.5)}},
    xmin=5, xmax=250,
    ymin=0, ymax=6,
    xtick={2, 50, 100, 150, 200, 250},
    xticklabels={0, 50, 100, 150, 200, 250},
    axis line style={latex-latex},
    legend columns=1,
    legend style={text=black, at={(0.995, 0.12)}},
    legend image post style={black}
    ]
    


    \addplot [smooth, color=color_uav1, line width=1pt] table [y=min_mav_dist, x=time]{figure8_uav1.txt};
    \addplot [smooth, color=color_uav2, line width=1pt] table [y=min_mav_dist, x=time]{figure8_uav2.txt};
    \addplot [smooth, color=color_uav3, line width=1pt] table [y=min_mav_dist, x=time]{figure8_uav3.txt};
    \addplot [smooth, color=color_uav4, line width=1pt] table [y=min_mav_dist, x=time]{figure8_uav4.txt};
    \addplot [smooth, color=color_uav5, line width=1pt] table [y=min_mav_dist, x=time]{figure8_uav5.txt};
    \addplot [smooth, color=color_uav6, line width=1pt] table [y=min_mav_dist, x=time]{figure8_uav6.txt};
    \addplot [smooth, color=color_uav7, line width=1pt] table [y=min_mav_dist, x=time]{figure8_uav7.txt};
    \addplot [smooth, color=color_uav8, line width=1pt] table [y=min_mav_dist, x=time]{figure8_uav8.txt};
    \addplot [smooth, color=color_uav9, line width=1pt] table [y=min_mav_dist, x=time]{figure8_uav9.txt};

  \end{axis}
\end{tikzpicture}
    }\\\vspace*{-1.0em}
    \subfloat[Minimal distance to the closest obstacle.]{
      \begin{tikzpicture}[font=\scriptsize]
  \definecolor{color_uav1}{HTML}{1f77b4}
  \definecolor{color_uav2}{HTML}{ff7f0e}
  \definecolor{color_uav3}{HTML}{2ca02c}
  \definecolor{color_uav4}{HTML}{d62728}
  \definecolor{color_uav5}{HTML}{8c564b}
  \definecolor{color_uav6}{HTML}{e377c2}
  \definecolor{color_uav7}{HTML}{7f7f7f}
  \definecolor{color_uav8}{HTML}{bcbd22}
  \definecolor{color_uav9}{HTML}{17becf}

  \begin{axis}[ 
    width=0.52\textwidth,
    height=0.2\textwidth,
    grid=major,
    grid style={draw=gray!30,line width=.1pt},
    xlabel=time (s),
    ylabel=distance (m),
    xlabel style={
    yshift=1.4ex},
    y label style={at={(-0.04,0.5)}},
    xmin=5, xmax=250,
    ymin=0, ymax=6,
    xtick={2, 50, 100, 150, 200, 250},
    xticklabels={0, 50, 100, 150, 200, 250},
    ytick={0, 1, 2, 4, 6},
    yticklabels={0, 1, 2, 4, 6},
    axis line style={latex-latex},
    legend columns=1,
    legend style={text=black, at={(0.995, 0.12)}},
    legend image post style={black}
    ]
    


    \addplot [smooth, color=color_uav1, line width=1pt] table [y=cl_obst, x=time]{figure8_uav1.txt};
    \addplot [smooth, color=color_uav2, line width=1pt] table [y=cl_obst, x=time]{figure8_uav2.txt};
    \addplot [smooth, color=color_uav3, line width=1pt] table [y=cl_obst, x=time]{figure8_uav3.txt};
    \addplot [smooth, color=color_uav4, line width=1pt] table [y=cl_obst, x=time]{figure8_uav4.txt};
    \addplot [smooth, color=color_uav5, line width=1pt] table [y=cl_obst, x=time]{figure8_uav5.txt};
    \addplot [smooth, color=color_uav6, line width=1pt] table [y=cl_obst, x=time]{figure8_uav6.txt};
    \addplot [smooth, color=color_uav7, line width=1pt] table [y=cl_obst, x=time]{figure8_uav7.txt};
    \addplot [smooth, color=color_uav8, line width=1pt] table [y=cl_obst, x=time]{figure8_uav8.txt};
    \addplot [smooth, color=color_uav9, line width=1pt] table [y=cl_obst, x=time]{figure8_uav9.txt};

  \end{axis}
\end{tikzpicture}
    }
    \caption{
      Navigation of a decentralized swarm of 9 homogeneous \acp{UAV} in a forest-like environment with a high density of circular obstacles -- tree trunks (a).
      The experiment showcases the cooperative steering within the environment and the emerging properties of mutual long-term cohesion (b), safe mutual separation (c), and reliable obstacle avoidance (d).
}
    \label{fig:sim_9mavs_in_forest}
  \end{figure}

\begin{figure}[t]
    \centering
    \begingroup
    \captionsetup{width=0.96\columnwidth}
    \subfloat[Swarm of 3 \acp{UAV} navigating through an artificial forest.]{
    \input{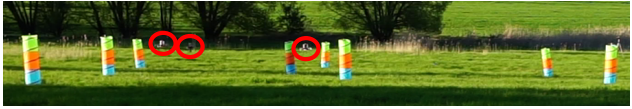}
    }
    \\
    \subfloat[Onboard RGB view from one of the homogeneous units.]{
    \input{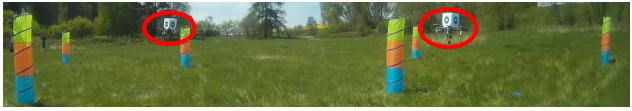}
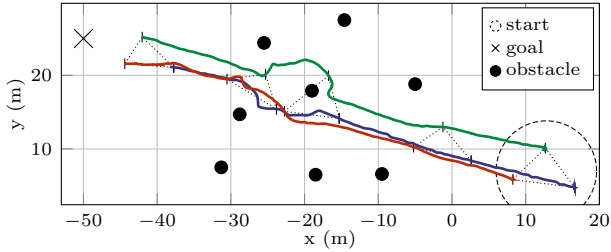
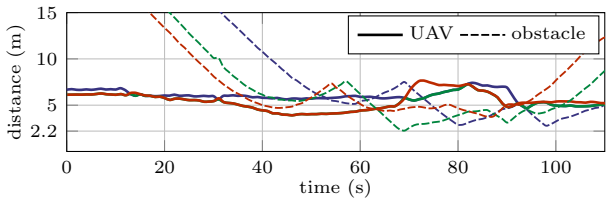
    }
    \\
    \subfloat[
    Trajectories of individual \acp{UAV} (coded by color).
    The dotted triangles represent the swarm constellation ($\alpha$-lattices) at a given time, which highlights the compactness and the flexibility of the swarm navigating amidst obstacles.
    ]{
    \begin{tikzpicture}[font=\scriptsize]

  \pgfplotsset{select coords between index/.style 2 args={
    x filter/.code={
        \ifnum\coordindex<#1\def\pgfmathresult{}\fi
        \ifnum\coordindex>#2\def\pgfmathresult{}\fi
    }
}}
 
  \definecolor{color_uav1}{rgb}{0.22, 0.2, 0.502}
  \definecolor{color_uav2}{rgb}{0, .522, .243}
  \definecolor{color_uav3}{rgb}{0.737,0.165,0}

  \pgfplotstableread[col sep=space]{figure9_lattices.txt}{\tablelattices}

  \begin{axis}[ 
    width=1.05\columnwidth,
    grid=major,
    xlabel=x (m),
    ylabel=y (m),
    xlabel style={
    yshift=1.4ex},
    y label style={at={(-0.05,0.5)}},
    xmin=-53, xmax=20,
    axis equal image,
    legend columns=1,
    legend style={text=black, row sep=-0.4ex, at={(0.99, 0.98)}},
    legend entries={start, goal, obstacle},
    legend cell align={left},
    clip marker paths=true,
    ]
    \addlegendimage{only marks,mark=o, dash pattern=on 1pt off 1pt on 1pt off 1pt}
    \addlegendimage{only marks,mark=x}
    \addlegendimage{only marks,mark=*,mark options={solid}}

    \addplot [mark=o, color=black, mark options={scale=9.5}, dash pattern=on 2pt off 1pt on 2pt off 1pt] coordinates {(12.8, 7.0)};
    \addplot [mark=x, color=black, mark options={scale=2.5}] coordinates {(-50, 25)};
          
    \addplot [densely dotted, mark=|, select coords between index={0}{3}, mark options={solid}] table {\tablelattices};
    \addplot [densely dotted, mark=|, select coords between index={4}{7}, mark options={solid}] table {\tablelattices};
    \addplot [densely dotted, mark=|, select coords between index={8}{11}, mark options={solid}] table {\tablelattices};
    \addplot [densely dotted, mark=|, select coords between index={12}{15}, mark options={solid}] table {\tablelattices};
    \addplot [densely dotted, mark=|, select coords between index={16}{19}, mark options={solid}] table {\tablelattices};

    \addplot [smooth, color=color_uav1, line width=1pt] table [y=y, x=x]{figure9_uav1.txt};
    \addplot [smooth, color=color_uav2, line width=1pt] table [y=y, x=x]{figure9_uav2.txt};
    \addplot [smooth, color=color_uav3, line width=1pt] table [y=y, x=x]{figure9_uav3.txt};
    \addplot+[only marks, mark=*, color=black, mark options={fill=black, scale=1.2}] table [y=y, x=x]{figure9_obstacles.txt};

  \end{axis}
\end{tikzpicture}
    }
    \vspace*{-0.6em}
    \\
    \subfloat[
    Euclidean distance to the nearest \ac{UAV} and obstacle for each swarm agent (coded by color).
    The minimum distance reached is \SI{2.18}{\metre}.
    ]{
    \begin{tikzpicture}[font=\scriptsize]
  \definecolor{color_uav1}{rgb}{0.22, 0.2, 0.502}
  \definecolor{color_uav2}{rgb}{0, .522, .243}
  \definecolor{color_uav3}{rgb}{0.737,0.165,0}
  \begin{axis}[ 
    width=1.05\columnwidth,
    height=0.2\textwidth,
    grid=major,
    xlabel=time (s),
    ylabel=distance (m),
    xlabel style={
    yshift=1.0ex},
    y label style={at={(-0.06,0.5)}},
    xmin=0, xmax=110, ymin=0, ymax=15,
    ytick={2.2, 5, 10, 15},
    legend columns=2,
    legend style={text=black, at={(0.99, 0.98)}},
    legend image post style={black}
    ]
    \addplot [smooth, color=color_uav1, line width=1pt] table [y=min_mav_dist, x=time]{figure9_uav1.txt};
    \addplot [smooth, dash pattern=on 3pt off 1pt on 3pt off 1pt, color=color_uav1, line width=0.7pt] table [y=cl_obst, x=time]{figure9_uav1.txt};
    \addlegendentry{UAV}
    \addlegendentry{obstacle}

    \addplot [smooth, color=color_uav2, line width=1pt] table [y=min_mav_dist, x=time]{figure9_uav2.txt};
    \addplot [smooth, color=color_uav3, line width=1pt] table [y=min_mav_dist, x=time]{figure9_uav3.txt};

    \addplot [smooth, dash pattern=on 3pt off 1pt on 3pt off 1pt, color=color_uav2, line width=0.7pt] table [y=cl_obst, x=time]{figure9_uav2.txt};
    \addplot [smooth, dash pattern=on 3pt off 1pt on 3pt off 1pt, color=color_uav3, line width=0.7pt] table [y=cl_obst, x=time]{figure9_uav3.txt};

  \end{axis}
\end{tikzpicture}
    }
    \endgroup
    \caption{
      Aerial swarm of 3 homogeneous \acp{UAV} in a real-world forest-like environment filled with artificial obstacles.
    }
    \label{fig:exp_louka}
  \end{figure}

  Second, an aerial swarm of 3 \acp{UAV} was experimentally deployed in a real-world forest-like environment similar to \autoref{fig:sim_9mavs_in_forest}, in order to verify the abilities of the fully-decentralized swarming model to stabilize a set of \acp{UAV} in a decentralized manner, provide self-organizing behavior, and to navigate through an obstacle-filled environment.
  As explicitly shown in \autoref{fig:exp_louka}, even such a simplistic swarming model with only local information yields collision-free navigation (the minimum distance to an obstacle or to another \ac{UAV} was \SI{2.2}{\metre}) throughout the environment, and self-organizing compactness of the whole swarm during the entire flight.
  The experiment likewise shows the ability of the model to divide the group when overcoming an obstacle and to unite back again afterwards.
  This level of flexibility is important for fast and safe navigation within more complex environments in order to maximize the motion effectiveness.
  The flexibility is highlighted by dotted triangles, which represent the geometric configuration of the swarm in time.
  Let us call this flock geometry an $\alpha$-lattice according to~\cite{1605401} and use it to represent a self-organizing structure, where individual inter-particle distances converge to a common value.
    This geometric configuration allows for small deviations from the expected structure (especially for particles in an environment with obstacles), which can be further quantified by \textit{deviation energy} and can be used to evaluate the swarming model convergence.
    The deviation energy is derived in~\cite{1605401} and represents a non-smooth potential function of a set of particles, where the $\alpha$-lattice configuration lies at its global minimum.




  \subsection{UVDAR in Control Feedback}
  \label{ssec:exp_uvdar}
  

  To verify the feasibility of the complete system defined in \autoref{fig:system_architecture}, \ac{UVDAR} vision-based mutual relative localization is deployed in the position control feedback loop of each homogeneous swarm agent.
  \hl{Throughout the experiment, the individual }\acp{UAV}\hl{ employ }\ac{GNSS}\hl{ for self-state estimation.
  This is required to stabilize the flight of each dynamically unstable }\ac{UAV}\hl{ mid-flight in a large open-space, where the swarm was deployed.}
  However, the agents do not share any information through a communication network and instead they directly perceive the neighboring particles using \ac{UVDAR}.
  \hl{%
    The blinking frequencies of the {\acp{UAV}} (IDs) within the experiment were static and unique.
    This improves the performance of the UVDAR localization as unique IDs in the image stream help to separate occluded detections and track the units in time.
  }
To the best of our knowledge, this is the first deployment of a fully decentralized aerial swarming system in a real environment (outside laboratory-like conditions) with direct localization and with no communication or position sharing allowed.

\begin{figure}[t]
    \begin{center}
      \begingroup
      \captionsetup{width=1.0\columnwidth}
      \subfloat[%
      Aerial view on the decentralized swarm of 4 \acp{UAV} (red) and a static reference to assist with the scale perception (blue).
      ]{
    \input{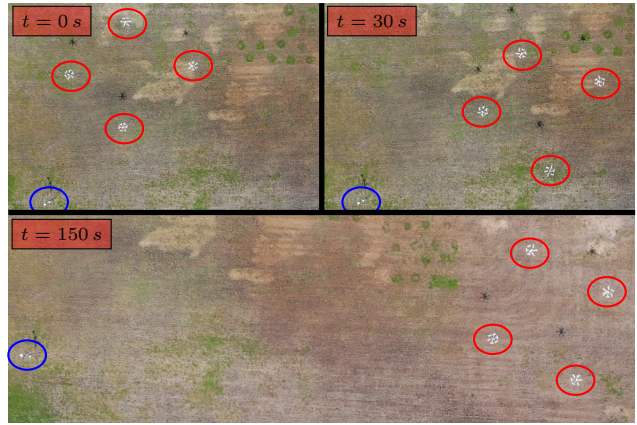}
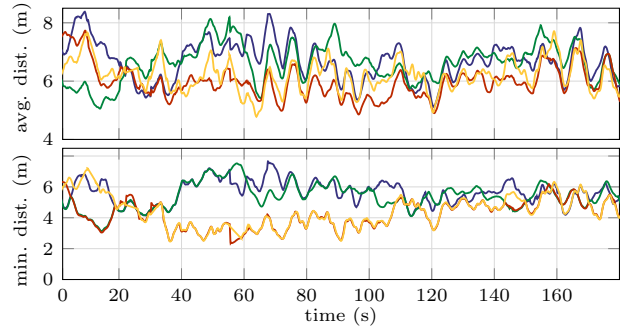
      }
      \\
      \subfloat[
        Average and minimal Euclidean distance among the homogeneous agents.
        The minimum distance reached is \SI{2.04}{\metre}.
      ]{
    \begin{tikzpicture}[font=\scriptsize]
  \definecolor{color_uav2}{rgb}{0.22, 0.2, 0.502}
  \definecolor{color_uav3}{rgb}{0, .522, .243}
  \definecolor{color_uav4}{rgb}{0.737,0.165,0}
  \definecolor{color_uav5}{rgb}{1.0,0.79,0.23}

  \pgfplotstableread[col sep=space]{figure10b_uav2.txt}{\tableuavtwo}
  \pgfplotstableread[col sep=space]{figure10b_uav3.txt}{\tableuavthree}
  \pgfplotstableread[col sep=space]{figure10b_uav4.txt}{\tableuavfour}
  \pgfplotstableread[col sep=space]{figure10b_uav5.txt}{\tableuavfive}
  \begin{axis}[ 
    name=avg,
    width=1.08\columnwidth,
    height=0.4\columnwidth,
    grid=major,
    grid style={draw=gray!30,line width=.1pt},
    ylabel=avg. dist. (m),
    y label style={at={(-0.04,0.55)}},
    xmin=2, xmax=180, ymin=4, ymax=8.5,
    xtick={2, 20, 40, 60, 80, 100, 120, 140, 160},
    xticklabels={},
    legend columns=2,
    legend style={text=black, at={(0.995, 0.12)}},
    legend image post style={black}
    axis line style={latex-latex},
    ]
    \addplot [smooth, color=color_uav2, line width=0.7pt] table [y=avg_mav_dist, x=time]{\tableuavtwo};
    \addplot [smooth, color=color_uav3, line width=0.7pt] table [y=avg_mav_dist, x=time]{\tableuavthree};
    \addplot [smooth, color=color_uav4, line width=0.7pt] table [y=avg_mav_dist, x=time]{\tableuavfour};
    \addplot [smooth, color=color_uav5, line width=0.7pt] table [y=avg_mav_dist, x=time]{\tableuavfive};
  \end{axis}

  \begin{axis}[ 
    name=min
    at=(avg.below south west), anchor=above north west,
    width=1.08\columnwidth,
    height=0.4\columnwidth,
    grid=major,
    grid style={draw=gray!30,line width=.1pt},
    xlabel=time (s),
    ylabel=min. dist. (m),
    xlabel style={
    yshift=1.0ex},
    y label style={at={(-0.04,0.45)}},
    xmin=2, xmax=180, ymin=0, ymax=8.5,
    xtick={2, 20, 40, 60, 80, 100, 120, 140, 160},
    xticklabels={0, 20, 40, 60, 80, 100, 120, 140, 160},
    ytick={0, 2, 4, 6, 8},
    yticklabels={0, 2, 4, 6, },
    legend columns=2,
    legend style={text=black, at={(0.995, 0.12)}},
    legend image post style={black}
    axis line style={latex-latex},
    yshift=-0.2em,
    ]
    \addplot [smooth, color=color_uav2, line width=0.7pt] table [y=min_mav_dist, x=time]{\tableuavtwo};
    \addplot [smooth, color=color_uav3, line width=0.7pt] table [y=min_mav_dist, x=time]{\tableuavthree};
    \addplot [smooth, color=color_uav4, line width=0.7pt] table [y=min_mav_dist, x=time]{\tableuavfour};
    \addplot [smooth, color=color_uav5, line width=0.7pt] table [y=min_mav_dist, x=time]{\tableuavfive};
  \end{axis}
\end{tikzpicture}
      }
      \endgroup
      \caption{
        A fully-decentralized swarm of 4 homogeneous \acp{UAV} navigating through an obstacle-less environment with \ac{UVDAR} integrated into the position control feedback, as outlined by the scheme in \autoref{fig:system_architecture}.
      }
      \label{fig:exp_cisar}
    \end{center}
  \end{figure}

  As explicitly shown in \autoref{fig:exp_cisar}, use of a local sensing method maintains the abilities of the bio-inspired swarming model, namely self-organizing behavior, together with collision-free and cohered navigation.
  The swarm is capable of navigation throughout the environment in a compact structural constellation without any external interference to a sequence of global navigation goals.
  The figure shows the ability to preserve a compact structure emerging from local \ac{UVDAR}-based perception (\autoref{fig:motivation_title_b} and \autoref{fig:uvdar_raw} show the perceived data of a single swarm agent in this particular experiment) and the elementary rules presented in \sref{sec:swarming}, while the homogeneous units do not share any information among themselves.


  \subsection{Analysis on Direct Observation Accuracy}
  \label{sec:direct_observation_accuracy}
  In real-world conditions, all estimation subsystems are incorporated with various measurements containing a stochastic noise element.
  The origin of this stochastic part is of numerous types (e.g., vibrations, discretization, approximations, sensor non-linearity, time desynchronization, lack of motion compensation, or optical discrepancies) and most of these inaccuracies need to be accounted for.
  For example, the stabilization and control system of \acp{UAV} requires a continuous stream of inertial measurements to cope with hardware-based and synchronization inaccuracies, in order to stabilize the dynamically unstable system in mid-flight.
  The influence of these inaccuracies needs to be carefully analyzed, and the results of the analyses must be incorporated into the design of a swarming model in order to compensate for the uncertainties of real-world systems. 

  As discussed in the review of the related literature (see \sref{sec:sota}), dense robotic swarms candidly communicate either external positioning estimates or individual global state estimates amid the swarm units.
  In addition to the requirements of the communication infrastructure, this methodology imitates the bio-inspired design of mutual localization by establishing the relative relations from the global data.
  This incorporates the global self-localization error, and can lead to dangerous decision making, and also to communication-based failures.
  However, our approach imitates biological systems by relying solely on direct localization without the need for known global states of the neighbors or of the unit itself.
  This bounds the overall performance of the system solely to the accuracy of the direct localization.
  It entirely removes the need for a communication infrastructure, and allows for full decentralization of the system architecture. 

  To analyze the impact of direct localization accuracy on the overall performance of our swarming framework, we present two inquiries: the influence of the error degree on the stability of a decentralized swarm, and the data-based accuracy of \ac{UVDAR} in real-world conditions.
  As our focus applies to vision-based direct localization, the error of 3D relative localization can be expressed in spherical coordinates -- radial distance, azimuth, and elevation -- separately.
  Bear in mind that due to the vision-based nature of \ac{UVDAR} discussed in \sref{sec:uvdar}, the statistical characteristics of the elevation error are assumed to be identical with the azimuthal error.
  To maintain simplicity, the elevation error is therefore omitted from the presentation of the results.
  
    \begin{figure}[t]
    \centering
    \begin{tikzpicture}[font=\scriptsize]

  \definecolor{color_uav2}{rgb}{0.22, 0.2, 0.502}
  \definecolor{color_uav3}{rgb}{0, .522, .243}
  \definecolor{color_uav4}{rgb}{0.737,0.165,0}
  \definecolor{color_uav5}{rgb}{1.0,0.79,0.23}

  \pgfplotstableread[col sep=space]{figure11_data.txt}{\tabledata}
  \begin{axis}[ 
    width=0.7\columnwidth,
    height=0.3\columnwidth,
    grid=major,
    grid style={draw=gray!30,line width=.1pt},
    xmin=0, xmax=4.5,
    scale only axis,
    xlabel=mutual distance (m),
    axis y line*=left,
    ylabel=$\sigma_r$ (m),
    ytick={0, 1, 2, 3, 4},
    yticklabels={0, 1, 2, 3, 4},
    legend columns=2,
    legend style={cells={align=left}, text=black, column sep=0.1cm, at={(1.0, 1.21)}, font=\scriptsize},
    axis line style={-},
    ]
    \addlegendentry{Minimal}
    \addlegendimage{only marks, mark=o, color=color_uav4, ultra thick}
    \addlegendentry{Average}
    \addlegendimage{only marks, mark=x, color=color_uav2, ultra thick}

    \addplot+[only marks, mark=x, color=color_uav2, line width=0.7pt, very thick] table [y=sigma_r, x=avg]{\tabledata};
  \end{axis}

  \begin{axis}[ 
    width=0.7\columnwidth,
    height=0.3\columnwidth,
    scale only axis,
    xmin=0, xmax=4.5,
    hide x axis,
    axis x line=none,
    axis y line*=right,
    ytick={0.04, 0.1, 0.16, 0.22, 0.28},
    yticklabels={0.04, 0.10, 0.16, 0.22, 0.28},
    ylabel=$\sigma_{\Psi}$ (rad),
    ]
    \addplot+[only marks, mark=o, color=color_uav4, line width=0.7pt, very thick] table [y=sigma_az, x=min]{\tabledata};
  \end{axis}

  \end{tikzpicture}
    \caption{
      Dependency of the direct localization accuracy on the stability of an aerial swarm.
      \hl{%
        The plot shows exponential decline of the minimum distance amid the swarm units with a growing degree of the localization error.
      The localization error is modeled as a multivariate normal distribution with uncorrelated zero-mean variables: the radial distance (standard deviation {$\sigma_r$}) and the relative azimuth (standard deviation {$\sigma_{\Psi}$}).
      }
    }
    \label{fig:analysis_on_accuracy}
  \end{figure}
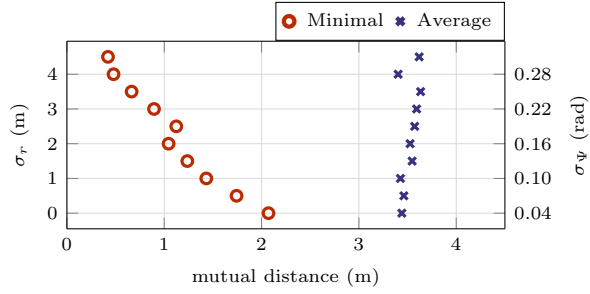
  

  The impact of a direct localization error on the stability of a swarm was analyzed on a set of computational simulations.
  A decentralized swarm of \acp{UAV} with simulated dynamics, control \& state estimation disturbances, and sensory inaccuracies, was deployed in scenarios with various degrees of the direct localization error according to \autoref{fig:analysis_on_accuracy}.
  Although the data show the minimum influence of the error on the average distance among the swarm units, the stochastic element induces oscillations of the mutual distances.
  These deviations from a consensual mutual distance arise directly from the inaccuracy of direct localization and from time-based and dynamics-based delays.
  This has a negative impact on the stability properties of the entire swarm, as shown by the exponential decline of the minimal distance amid the swarm units with the increasing degree of the radial distance and the relative azimuth error in \autoref{fig:analysis_on_accuracy}.
  In real-world systems, a suitable swarm density must be thoroughly considered \hl{with respect to} the accuracy and the reliability of the direct localization in order to prevent undesired collisions.



  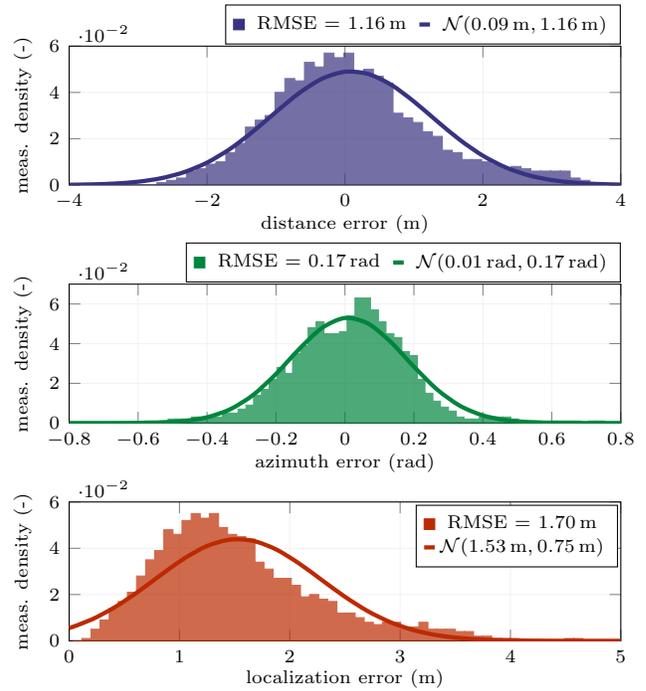
\begin{figure}[t]
    \begin{center}
    \begin{tikzpicture}[font=\scriptsize]

  \definecolor{color_uav2}{rgb}{0.22, 0.2, 0.502}
  \definecolor{color_uav3}{rgb}{0, .522, .243}
  \definecolor{color_uav4}{rgb}{0.737,0.165,0}
  \definecolor{color_uav5}{rgb}{1.0,0.79,0.23}

  \makeatletter
  \newcommand{\pgfplotsxmin}{\pgfplots@xmin}
  \newcommand{\pgfplotsxmax}{\pgfplots@xmax}
  \makeatother

  \newcommand\gauss[3]{1/(#2*sqrt(2*pi))*exp(-((x-#1)^2)/(2*#2^2))/#3} 
  \pgfplotstableread[col sep=space]{figure12_uvdar_accuracy.txt}{\tablegauss}
  \pgfplotstableread[col sep=space]{figure12_uvdar_histograms.txt}{\tabledata}

  \pgfplotstablegetelem{0}{rmse_r}\of\tablegauss
  \pgfmathsetmacro{\rmser}{\pgfplotsretval}
  \pgfplotstablegetelem{0}{mean_r}\of\tablegauss
  \pgfmathsetmacro{\meanr}{\pgfplotsretval}
  \pgfplotstablegetelem{0}{std_r}\of\tablegauss
  \pgfmathsetmacro{\stdr}{\pgfplotsretval}
  \pgfplotstablegetelem{0}{norm_r}\of\tablegauss
  \pgfmathsetmacro{\normr}{\pgfplotsretval}

  \pgfplotstablegetelem{0}{rmse_yaw}\of\tablegauss
  \pgfmathsetmacro{\rmseyaw}{\pgfplotsretval}
  \pgfplotstablegetelem{0}{mean_yaw}\of\tablegauss
  \pgfmathsetmacro{\meanyaw}{\pgfplotsretval}
  \pgfplotstablegetelem{0}{std_yaw}\of\tablegauss
  \pgfmathsetmacro{\stdyaw}{\pgfplotsretval}
  \pgfplotstablegetelem{0}{norm_yaw}\of\tablegauss
  \pgfmathsetmacro{\normyaw}{\pgfplotsretval}

  \pgfplotstablegetelem{0}{rmse_dist}\of\tablegauss
  \pgfmathsetmacro{\rmsed}{\pgfplotsretval}
  \pgfplotstablegetelem{0}{mean_dist}\of\tablegauss
  \pgfmathsetmacro{\meand}{\pgfplotsretval}
  \pgfplotstablegetelem{0}{std_dist}\of\tablegauss
  \pgfmathsetmacro{\stdd}{\pgfplotsretval}
  \pgfplotstablegetelem{0}{norm_dist}\of\tablegauss
  \pgfmathsetmacro{\normd}{\pgfplotsretval}

  \begin{axis}[ 
    name=plot_dist,
    width=0.52\textwidth,
    height=0.2\textwidth,
    grid=both,
    grid style={draw=gray!10,line width=.1pt},
    xlabel=distance error (m),
    ylabel=meas. density (-),
    xlabel style={
    yshift=1.0ex},
    y label style={at={(-0.05,0.5)}},
    xmin=-4, xmax=4, ymin=0, ymax=0.06,
    legend columns=2,
    legend style={cells={align=left}, text=black, column sep=0.1cm, at={(1.0, 1.3)}, font=\scriptsize},
    axis line style={latex-latex},
    ]
    \addlegendentry{RMSE = $\SI{\rmser}{\metre}$}
    \addlegendimage{only marks, mark=square*, color=color_uav2}
    \addlegendentry{$\mathcal{N}(\SI{\meanr}{\metre}, \SI{\stdr}{\metre})$}
    \addlegendimage{only marks, mark=-, color=color_uav2, ultra thick}

    \addplot+[no markers, ybar interval, fill, color=color_uav2, fill opacity=0.5, opacity=0.7] table [x=bins_r, y=data_r] {\tabledata};
    \addplot+[no markers, ultra thick, color=color_uav2, domain=\pgfplotsxmin:\pgfplotsxmax, smooth] {\gauss{\meanr}{\stdr}{\normr}};

    \end{axis}
  
    \begin{axis}[ 
    name=plot_azimuth,
    at=(plot_dist.below south west), anchor=above north west,
    width=0.52\textwidth,
    height=0.2\textwidth,
    grid=both,
    grid style={draw=gray!10,line width=.1pt},
    xlabel=azimuth error (rad),
    ylabel=meas. density (-),
    xlabel style={
    yshift=1.0ex},
    y label style={at={(-0.05,0.5)}},
    xmin=-0.8, xmax=0.8, ymin=0, ymax=0.07,
    legend columns=2,
    legend style={cells={align=left}, text=black, column sep=0.1cm, at={(1.0, 1.3)}, font=\scriptsize},
    axis line style={latex-latex},
    ]
    \addlegendentry{RMSE = $\SI{\rmseyaw}{\radian}$}
    \addlegendimage{only marks, mark=square*, color=color_uav3}
    \addlegendentry{$\mathcal{N}(\SI{\meanyaw}{\radian}, \SI{\stdyaw}{\radian})$}
    \addlegendimage{only marks, mark=-, color=color_uav3, ultra thick}

    \addplot+[no markers, ybar interval, fill, color=color_uav3, fill opacity=0.5, opacity=0.7] table [x=bins_yaw, y=data_yaw] {\tabledata};
    \addplot+[no markers, ultra thick, color=color_uav3, domain=\pgfplotsxmin:\pgfplotsxmax, smooth] {\gauss{\meanyaw}{\stdyaw}{\normyaw}};

    \end{axis}

    \begin{axis}[ 
    name=plot_d,
    at=(plot_azimuth.below south west), anchor=above north west,
    width=0.52\textwidth,
    height=0.2\textwidth,
    grid=both,
    grid style={draw=gray!10,line width=.1pt},
    xlabel=localization error (m),
    ylabel=meas. density (-),
    xlabel style={
    yshift=1.0ex},
    y label style={at={(-0.05,0.5)}},
    xmin=0, xmax=5, ymin=0, ymax=0.06,
    legend columns=1,
    legend style={cells={align=left}, text=black, row sep=-0.4ex, at={(0.995, 0.98)}, font=\scriptsize},
    axis line style={latex-latex},
    ]
    \addlegendentry{RMSE = $\SI{\rmsed}{\metre}$}
    \addlegendimage{only marks, mark=square*, color=color_uav4}
    \addlegendentry{$\mathcal{N}(\SI{\meand}{\metre}, \SI{\stdd}{\metre})$}
    \addlegendimage{only marks, mark=-, color=color_uav4, ultra thick}

    \addplot+[no markers, ybar interval, fill, color=color_uav4, fill opacity=0.5, opacity=0.7] table [x=bins_d, y=data_d] {\tabledata};
    \addplot+[no markers, ultra thick, color=color_uav4, domain=\pgfplotsxmin:\pgfplotsxmax, smooth] {\gauss{\meand}{\stdd}{\normd}};

    \end{axis}
\end{tikzpicture}
      \caption{
        Quantitative accuracy of \ac{UVDAR} direct localization \hl{with respect to} \ac{GNSS} positioning in real-world conditions. 
        The figure shows error histograms and their normalized normal distribution $\mathcal{N}(\mu, \sigma)$ approximations of the directly estimated relative distance, the relative azimuth, and the global 3D position.
      }
      \label{fig:uvdar_accuracy}
    \end{center}
  \end{figure}


  The accuracy of \ac{UVDAR} in real-world conditions during the deployment of the decentralized swarm of 4 \ac{UAV} units in an open environment (see \autoref{fig:exp_cisar}) is expressed by the error histograms in \autoref{fig:uvdar_accuracy}.
  During this experiment, the self-localization of the individual \acp{UAV} was arranged by \ac{GNSS}.
  The statistical analysis uses global positioning for a quantitative evaluation of the direct localization accuracy.
  Although global positioning yields a relatively high error, the state estimation module (see \sref{sec:system_architecture}) fuses this global state estimate with inertial measurements, which makes the output estimate robust towards sudden short-term changes.
  The positioning is still prone to long-term drift, which is minimal in terms of \ac{GNSS} and therefore does not significantly impact the evaluation of the direct localization within a dense swarm.
  The fused global estimate is therefore used as ground truth data for the quantitative evaluation in \autoref{fig:uvdar_accuracy}.
  This evaluation on real-world data shows the ability of \ac{UVDAR} to estimate the relative distance with \SI{1.16}{\metre} \ac{RMSE} and the relative azimuth with \ac{RMSE} of \SI{0.17}{\radian}.
  These separated errors then combine together with the elevation estimate to anticipate the relative 3D position of the neighboring particles within a moving aerial swarm with \ac{RMSE} of \SI{1.7}{\metre}.

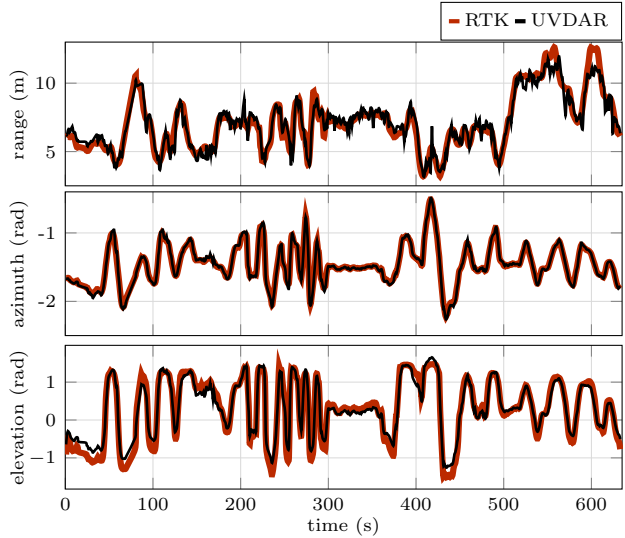
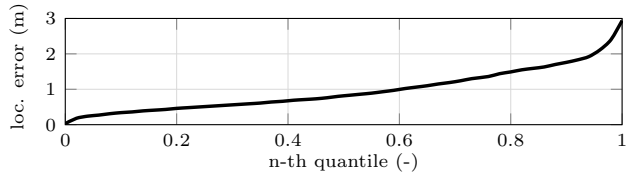
\begin{figure}[t]
    \centering
    \begingroup
    \captionsetup{width=0.95\columnwidth}
    \subfloat[Relative localization represented by the spherical coordinates (expressed in the origin of the camera)]{
      \begin{tikzpicture}[font=\scriptsize]

  \definecolor{color_uvdar}{rgb}{0,0,0}
  \definecolor{color_rtk}{rgb}{0.737,0.165,0}
  \pgfplotstableread[col sep=space]{figure13_data.txt}{\tabledata}

  \begin{axis}[ 
    name=range,
    width=1.08\columnwidth,
    height=0.42\columnwidth,
    grid=major,
    grid style={draw=gray!30,line width=.1pt},
    ylabel=range (m),
    y label style={at={(-0.05,0.5)}},
    xmin=0, xmax=635, ymin=2.5, ymax=13,
    xticklabels={},
    legend columns=2,
    legend style={at={(1.0, 1.28)}},
    axis line style={latex-latex},
    ]

    \addlegendimage{only marks, mark=-, color=color_rtk, ultra thick}
    \addlegendentry{RTK}
    \addlegendimage{only marks, mark=-, color=color_uvdar, ultra thick}
    \addlegendentry{UVDAR}

    \addlegendimage{only marks, mark=-, color=color_uav2, ultra thick}
    \addplot [smooth, color=color_rtk, line width=2.35pt] table [y=rtk_r, x=t]{\tabledata};
    \addplot [smooth, color=color_uvdar, line width=1.0pt] table [y=uvdar_r, x=t]{\tabledata};
  \end{axis}

  \begin{axis}[ 
    name=az,
    at=(range.below south west), anchor=above north west,
    width=1.08\columnwidth,
    height=0.42\columnwidth,
    grid=major,
    grid style={draw=gray!30,line width=.1pt},
    ylabel=azimuth (rad),
    y label style={at={(-0.05,0.5)}},
    xmin=0, xmax=635, ymin=-2.5, ymax=-0.4,
    ytick={-2, -1},
    yticklabels={-2, -1},
    xticklabels={},
    legend columns=2,
    axis line style={latex-latex},
    yshift=0.5em,
    ]


    \addplot [smooth, color=color_rtk, line width=2.35pt] table [y=rtk_az, x=t]{\tabledata};
    \addplot [smooth, color=color_uvdar, line width=1.0pt] table [y=uvdar_az, x=t]{\tabledata};
  \end{axis}

  \begin{axis}[ 
    name=el,
    at=(az.below south west), anchor=above north west,
    width=1.08\columnwidth,
    height=0.42\columnwidth,
    grid=major,
    grid style={draw=gray!30,line width=.1pt},
    xlabel=time (s),
    ylabel=elevation (rad),
    xlabel style={
    yshift=1.0ex},
    y label style={at={(-0.05,0.5)}},
    xmin=0, xmax=635, 
    legend columns=2,
    axis line style={latex-latex},
    yshift=0.5em,
    ]
    \addplot [smooth, color=color_rtk, line width=2.35pt] table [y=rtk_el, x=t]{\tabledata};
    \addplot [smooth, color=color_uvdar, line width=1.0pt] table [y=uvdar_el, x=t]{\tabledata};
  \end{axis}

\end{tikzpicture}
    }\\\vspace*{-0.7em}
    \subfloat[Quantiles of the absolute 3D localization error]{
      \begin{tikzpicture}[font=\scriptsize]
  \definecolor{color_q}{rgb}{0,0,0}
  \pgfplotstableread[col sep=space]{figure13_quantiles.txt}{\tabledata}

  \begin{axis}[ 
    name=range,
    width=1.08\columnwidth,
    height=0.36\columnwidth,
    grid=major,
    grid style={draw=gray!30,line width=.1pt},
    xlabel=n-th quantile (-),
    ylabel=loc. error (m),
    xlabel style={
    yshift=1.0ex},
    y label style={at={(-0.05,0.55)}},
    xmin=0, xmax=1, ymin=0, ymax=3,
    legend columns=2,
    legend style={at={(1.0, 1.31)}},
    axis line style={latex-latex},
    ]


    \addplot [smooth, color=color_q, line width=1.3pt] table [y=q_d, x=x]{\tabledata};
  \end{axis}

\end{tikzpicture}
    }
    \caption{%
      \hl{Real-world accuracy of} \ac{UVDAR}\hl{ direct localization in a controlled environment --- tracking of a single mid-flight }\ac{UAV}\hl{ relative to a static ground }\ac{UV}\hl{ camera.
      The }\ac{UVDAR}\hl{ localization is compared to ground-truth data obtained with the use of }\ac{RTK}-\ac{GNSS}.
      \hl{The absolute }\ac{RMSE}\hl{ of the relative 3D localization in this experiment reached }\SI{1.11}{\metre} \hl{(the median is }\SI{0.81}{\metre}\hl{).}%
    }
    \label{fig:uvdar_accuracy_rtk}
    \endgroup
  \end{figure}

  \hl{%
    The accuracy of }\ac{UVDAR}\hl{ in real-world conditions is further analyzed in a controlled outdoor environment.
  During an independent experiment, a position of a single mid-air }\ac{UAV}\hl{ was tracked in data from a static ground camera equipped with }\ac{UVDAR}\hl{ and was compared to a precise }\ac{RTK}-\ac{GNSS}\hl{ ({\SI{2}{\centi\metre}} accuracy) serving as a ground-truth.
  The comparison of the relative localization with the ground-truth data is shown in }\autoref{fig:uvdar_accuracy_rtk}\hl{.
  The data show the property of }\ac{UVDAR}\hl{ to localize an aerial unit with }\ac{RMSE}\hl{ of {\SI{1.11}{\metre}}.%
}
  
  The concluded accuracy is particularly important for the design of bio-inspired systems employing the \ac{UVDAR} sensor as a source of direct localization of neighboring units.
  The quantitative results of this analysis allow for appropriate compensation of the inaccuracies and credible verification of swarming models in a simulator, which necessarily precede real-world applications.
  


  \section{Conclusion}
  \label{sec:conclusion}
  This article has presented a framework for deploying fully-decentralized aerial swarms in real-world conditions with the use of vision-based \ac{UV} mutual relative localization of neighboring swarm units.
  The framework architecture, as well as the off-the-shelf \ac{UVDAR} system for direct localization within an aerial swarm, has been thoroughly discussed, has been deployed on a decentralized swarm of \acp{UAV} in real-world environments, and its performance has been analyzed.
  The experimental analysis verified the stability of \ac{UVDAR} as an input into a fully-decentralized swarming architecture, which embodies the communication-free and local-information swarming models that are commonly found among biological systems.
  The set of real-world experiments is, to the best of our knowledge, the first deployment of a decentralized swarm of \acp{UAV} with no use of a communication network or of external localization.
  The system is provided as open source, and is designed for simple integration and verification of flocking techniques (often bio-inspired), respecting the requirements of the swarming paradigm.

  \ack
  This work was supported 
  by the Czech Science Foundation (GA\v{C}R) under research project No. 20-10280S,
  by CTU grant no SGS20/174/OHK3/3T/13,
  by funding from the European Union's Horizon 2020 research and innovation programme under grant agreement No. 871479, and
  by OP VVV project CZ.02.1.01/0.0/0.0/16 019/0000765 "Research Center for Informatics".
  \hl{The authors thank Daniel He\v{r}t for preparing all the necessary equipment required for the experimental analysis.}

  \section*{Supplementary Materials}
  The multimedia materials supporting the article are available at \url{http://mrs.felk.cvut.cz/research/swarm-robotics}.
  The entire system is also available as open source at \url{https://github.com/ctu-mrs}.

  \section*{ORCID}
  Pavel Petr\'{a}\v{c}ek $\orcidicon{0000-0002-0887-9430}$: 0000-0002-0887-9430\hfill\\
  Viktor Walter $\orcidicon{0000-0001-8693-6261}$: 0000-0001-8693-6261\hfill\\
  Tom\'{a}\v{s} B\'{a}\v{c}a $\orcidicon{0000-0001-9649-8277}$: 0000-0001-9649-8277\hfill\\
  Martin Saska $\orcidicon{0000-0001-7106-3816}$: 0000-0001-7106-3816\hfill

  \begin{acronym}
    \acro{ML}[ML]{machine learning}
    \acro{CNN}[CNN]{Convolutional Neural Network}
    \acro{MAV}[MAV]{micro aerial vehicle}
    \acro{UAV}[UAV]{unmanned aerial vehicle}
    \acro{UV}[UV]{ultraviolet}
    \acro{IR}[IR]{infrared}
    \acro{IIR}[IR]{Infrared}
    \acro{UVDAR}[UVDAR]{UltraViolet Direction And Ranging}
    \acro{UT}[UT]{Unscented Transform}
    \acro{GNSS}[GNSS]{global navigation satellite system}
    \acro{RTK}[RTK]{real-time kinematic}
    \acro{MOCAP}[mo-cap]{Motion capture}
    \acro{ROS}[ROS]{Robot Operating System}
    \acro{MPC}[MPC]{model predictive control}
    \acro{UWB}[UWB]{Ultra-Wideband ranging}
    \acro{DBSCAN}[DBSCAN]{density-based spatial clustering of applications with noise}
    \acro{IMU}[IMU]{inertial measurement unit}
    \acro{EKF}[EKF]{Extended Kalman Filter}
    \acro{SLAM}[SLAM]{simultaneous localization and mapping}
    \acro{RMSE}[RMSE]{root mean square error}
    \acro{ICNIRP}[ICNIRP]{International Commission on Non-Ionizing Radiation Protection}
  \end{acronym}

  \section*{References}
  \bibliographystyle{iopart-num}
  \bibliography{main}

\providecommand{\newblock}{}
\begin{thebibliography}{10}
\expandafter\ifx\csname url\endcsname\relax
  \def\url#1{{\tt #1}}\fi
\expandafter\ifx\csname urlprefix\endcsname\relax\def\urlprefix{URL }\fi
\providecommand{\eprint}[2][]{\url{#2}}

\bibitem{9024023}
{\v{S}tibinger} P, {B\'{a}\v{c}a} T and {Saska} M 2020 {Localization of
  Ionizing Radiation Sources by Cooperating Micro Aerial Vehicles with Pixel
  Detectors in Real-Time} {\em IEEE Robot. Autom. Lett.\/} {\bf 5} 3634--3641

\bibitem{7989609}
{Gassner} M, {Cieslewski} T and {Scaramuzza} D 2017 {Dynamic collaboration
  without communication: Vision-based cable-suspended load transport with two
  quadrotors} {\em {Int. Conf. on Robotics and Automation}\/} pp 5196--5202

\bibitem{Trianni2008EvolutionarySR}
Trianni V 2008 {\em {Evolutionary Swarm Robotics: Evolving Self-Organising
  Behaviours in Groups of Autonomous Robots (Studies in Computational
  Intelligence)}\/} 1st ed (Springer Publishing Company, Incorporated)

\bibitem{sturnus_vulgaris}
Young G~F, Scardovi L, Cavagna A, Giardina I and Leonard N~E 2013 {Starling
  Flock Networks Manage Uncertainty in Consensus at Low Cost} {\em PLOS
  Computational Biology\/} {\bf 9} 1--7

\bibitem{swarm_overview}
{Chung} S, {Paranjape} A~A, {Dames} P, {Shen} S and {Kumar} V 2018 {A Survey on
  Aerial Swarm Robotics} {\em IEEE Trans. Robot.\/} {\bf 34} 837--855

\bibitem{uwb}
{Nguyen} T, {Qiu} Z, {Nguyen} T~H, {Cao} M and {Xie} L 2019 {Distance-Based
  Cooperative Relative Localization for Leader-Following Control of MAVs} {\em
  IEEE Robot. Autom. Lett.\/} {\bf 4} 3641--3648

\bibitem{bt_directcommunication}
Bhavana T, Nithya M and Rajesh M 2017 {Leader-follower co-ordination of
  multiple robots with obstacle avoidance} {\em SmartTechCon\/}

\bibitem{gpsdenied}
Saska M, B{\'a}{\v{c}}a T, Thomas J, Chudoba J, P\v{r}eu\v{c}il L, Krajn{\'i}k
  T, Faigl J, Loianno G and Kumar V 2017 {System for deployment of groups of
  unmanned micro aerial vehicles in GPS-denied environments using onboard
  visual relative localization} {\em Auton. Robot.\/} {\bf 41} 919--944

\bibitem{whycon}
Faigl J, Krajn{\'i}k T, Chudoba J, P\v{r}eu\v{c}il L and Saska M 2013 Low-cost
  embedded system for relative localization in robotic swarms {\em Int. Conf.
  on Robotics and Automation\/} pp 993--998

\bibitem{ir_us_indoor}
{De Silva} O, {Mann} G~K~I and {Gosine} R~G 2015 {An Ultrasonic and
  Vision-Based Relative Positioning Sensor for Multirobot Localization} {\em
  IEEE Sensors J.\/} {\bf 15} 1716--1726

\bibitem{ir_indoor}
{Yan} X, {Deng} H and {Quan} Q 2019 {Active Infrared Coded Target Design and
  Pose Estimation for Multiple Objects} {\em Int. Conf. on Intelligent Robots
  and Systems\/} pp 6885--6890

\bibitem{scaramuzza_blinkers}
Censi A, Strubel J, Brandli C, Delbruck T and Scaramuzza D 2013 {Low-latency
  localization by active LED markers tracking using a dynamic vision sensor}
  {\em Int. Conf. on Intelligent Robots and Systems\/} pp 891--898

\bibitem{lf_ir_2donly}
Park H, Choi I, Park S and Choi J 2013 {Leader-follower formation control using
  infrared camera with reflective tag} {\em 10th Int. Conf. on Ubiquitous
  Robots and Ambient Intelligence\/} pp 321--324

\bibitem{cnn}
Chaudhary K, Zhao M, Shi F, Chen X, Okada K and Inaba M 2017 {Robust real-time
  visual tracking using dual-frame deep comparison network integrated with
  correlation filters} {\em Int. Conf. on Intelligent Robots and Systems\/} pp
  6837--6842

\bibitem{8593405}
{Carrio} A, {Tordesillas} J, {Vemprala} S, {Saripalli} S, {Campoy} P and {How}
  J~P 2020 {Onboard Detection and Localization of Drones Using Depth Maps} {\em
  IEEE Access\/} {\bf 8} 30480--30490

\bibitem{intel}
Intel\textsuperscript{\textregistered} 2019 {Drones Light Up The Sky}
  \urlprefix\url{intel.com/content/www/us/en/technology-innovation/aerial-technology-light-show.html}

\bibitem{Vir_gh_2014}
Vir{\'{a}}gh C, V{\'{a}}s{\'{a}}rhelyi G, Tarcai N, Ször{\'{e}}nyi T, Somorjai
  G, Nepusz T and Vicsek T 2014 Flocking algorithm for autonomous flying robots
  {\em Bioinspir. Biomim.\/} {\bf 9} 025012

\bibitem{DBLP:journals/corr/VasarhelyiVSTSNV14}
V{\'{a}}s{\'{a}}rhelyi G, Vir{\'{a}}gh C, Somorjai G, Tarcai N,
  Sz{\"{o}}r{\'{e}}nyi T, Nepusz T and Vicsek T 2014 Outdoor flocking and
  formation flight with autonomous aerial robots {\em Int. Conf. on Intelligent
  Robots and Systems\/} pp 3866--3873

\bibitem{gabor2018}
V{\'a}s{\'a}rhelyi G, Vir{\'a}gh C, Somorjai G, Nepusz T, Eiben A~E and Vicsek
  T 2018 Optimized flocking of autonomous drones in confined environments {\em
  Sci. Robot.\/} {\bf 3}

\bibitem{ehang}
EHang 2019 {EHang Drone Formation Flight} \urlprefix\url{ehang.com/formation}

\bibitem{6095129}
{Hauert} S, {Leven} S, {Varga} M, {Ruini} F, {Cangelosi} A, {Zufferey} J and
  {Floreano} D 2011 {Reynolds flocking in reality with fixed-wing robots:
  Communication range vs. maximum turning rate} {\em Int. Conf. on Intelligent
  Robots and Systems\/} pp 5015--5020

\bibitem{Burkle2011}
B{\"u}rkle A, Segor F and Kollmann M 2011 {Towards Autonomous Micro UAV Swarms}
  {\em {J. Intell. Robot. Syst.}\/} {\bf 61} 339--353

\bibitem{Kushleyev2013}
Kushleyev A, Mellinger D, Powers C and Kumar V 2013 {Towards A Swarm of Agile
  Micro Quadrotors} {\em Auton. Robot.\/} {\bf 35} 287--300

\bibitem{8276634}
{Weinstein} A, {Cho} A, {Loianno} G and {Kumar} V 2018 {Visual Inertial
  Odometry Swarm: An Autonomous Swarm of Vision-Based Quadrotors} {\em IEEE
  Trans. Robot. Autom.\/} {\bf 3} 1801--1807

\bibitem{stirling}
Stirling T, Roberts J, Zufferey J~C and Floreano D 2012 Indoor navigation with
  a swarm of flying robots {\em Int. Conf. on Robotics and Automation\/}
  4641--4647

\bibitem{6942701}
{Nägeli} T, {Conte} C, {Domahidi} A, {Morari} M and {Hilliges} O 2014
  Environment-independent formation flight for micro aerial vehicles {\em Int.
  Conf. on Intelligent Robots and Systems\/} pp 1141--1146

\bibitem{Diggelen2015TheWF}
van Diggelen F and Enge P~K 2015 {The World’s first GPS MOOC and Worldwide
  Laboratory using Smartphones} {\em ION GNSS+\/} pp 361--369

\bibitem{7073497}
{Garcia Carrillo} L~R, {Fantoni} I, {Rondon} E and {Dzul} A 2015
  {Three-dimensional position and velocity regulation of a quad-rotorcraft
  using optical flow} {\em IEEE Trans. Aerosp. Electron. Syst.\/} {\bf 51}
  358--371

\bibitem{doi:10.1002/rob.21506}
Schmid K, Lutz P, Tomi\'{c} T, Mair E and Hirschmüller H 2014 {Autonomous
  Vision-based Micro Air Vehicle for Indoor and Outdoor Navigation} {\em J.
  Field Robot.\/} {\bf 31} 537--570

\bibitem{KohlbrecherMeyerStrykKlingaufFlexibleSlamSystem2011}
Kohlbrecher S, Meyer J, von Stryk O and Klingauf U 2011 {A Flexible and
  Scalable SLAM System with Full 3D Motion Estimation} {\em Int. Symp. on
  Safety, Security and Rescue Robotics\/} pp 155--160

\bibitem{beeclust}
Schmickl T and Hamann H 2016 {\em {BEECLUST: A swarm algorithm derived from
  honeybees: Derivation of the algorithm, analysis by mathematical models, and
  implementation on a robot swarm}\/} pp 95--137

\bibitem{BODI2015819}
Bodi M, Möslinger C, Thenius R and Schmickl" T 2015 {BEECLUST used for
  exploration tasks in Autonomous Underwater Vehicles} {\em 8th Int. Conf. on
  Mathematical Modelling\/} pp 819--824

\bibitem{8613876}
{Shah} D and {Vachhani} L 2019 {Swarm Aggregation Without Communication and
  Global Positioning} {\em IEEE Trans. Robot. Autom.\/} {\bf 4} 886--893

\bibitem{swarm_robots_review}
{Olaronke, Iroju and Ikono, Rhoda and Gambo, Ishaya and Ojerinde, Oluwaseun}
  2020 {A Systematic Review of Swarm Robots} {\em Current Journal of Applied
  Science and Technology\/} {\bf 39} 79--97

\bibitem{OH201783}
Oh H, Shirazi A~R, Sun C and Jin Y 2017 {Bio-inspired self-organising
  multi-robot pattern formation: A review} {\em Rob. Auton. Syst.\/} {\bf 91}
  83--100

\bibitem{Smith_2019}
Smith N~M, Dickerson A~K and Murphy D 2019 {Organismal aggregations exhibit
  fluidic behaviors: a review} {\em Bioinspir. Biomim.\/} {\bf 14} 031001

\bibitem{visualstabilization}
Saska M, Vakula J and P\v{r}eu\v{c}il L 2014 {Swarms of Micro Aerial Vehicles
  Stabilized Under a Visual Relative Localization} {\em {Int. Conf. on Robotics
  and Automation}\/} pp 3570--3575

\bibitem{uvdd2}
Walter V, Staub N, Saska M and Franchi A 2018 {Mutual Localization of UAVs
  based on Blinking Ultraviolet Markers and 3D Time-Position Hough Transform}
  {\em {14th Int. Conf. on Automation Science and Engineering}\/} pp 298--303

\bibitem{uvdar_ral}
{Walter} V, {Staub} N, {Franchi} A and {Saska} M 2019 {UVDAR System for Visual
  Relative Localization With Application to Leader–Follower Formations of
  Multirotor UAVs} {\em IEEE Trans. Robot. Autom.\/} {\bf 4} 2637--2644

\bibitem{uvdar_dataset_paper}
{Walter} V, {Vrba} M and {Saska} M 2020 {On training datasets for machine
  learning-based visual relative localization of micro-scale UAVs} {\em Int.
  Conf. on Robotics and Automation\/} Accepted

\bibitem{UV2004guidelines}
{International Commission on Non-Ionizing Radiation Protection and others} 2004
  {Guidelines on limits of exposure to ultraviolet radiation of wavelengths
  between 180 nm and 400 nm (incoherent optical radiation)} {\em Health
  Physics\/} {\bf 87} 171--186

\bibitem{led}
{ProLight Opto Technology Corporation} 2013 {ProLight PM2B-1LLE 1W UV Power LED
  Technical Datasheet}

\bibitem{Calovi_2014}
Calovi D~S, Lopez U, Ngo S, Sire C, Chat{\'{e}} H and Theraulaz G 2014
  {Swarming, schooling, milling: phase diagram of a data-driven fish school
  model} {\em New J. Phys.\/} {\bf 16} 015026

\bibitem{Reynolds:1987:FHS:37401.37406}
Reynolds C~W 1987 {Flocks, Herds and Schools: A Distributed Behavioral Model}
  {\em 14th Ann. Conf. on Computer Graphics and Interactive Techniques\/} pp
  25--34

\bibitem{1605401}
{Olfati-Saber} R 2006 Flocking for multi-agent dynamic systems: algorithms and
  theory {\em IEEE Trans. Autom. Control\/} {\bf 51} 401--420

\bibitem{Zhu2019DistributedMF}
Zhu H, Juhl J, Ferranti L and Alonso-Mora J 2019 {Distributed Multi-Robot
  Formation Splitting and Merging in Dynamic Environments} {\em Int. Conf. on
  Robotics and Automation\/}  9080--9086

\bibitem{DBLP:journals/corr/abs-1904-03742}
Erunsal I~K, Ventura R and Martinoli A 2019 {Nonlinear Model Predictive Control
  for 3D Formation of Multirotor Micro Aerial Vehicles with Relative Sensing in
  Local Coordinates} {\bf arXiv:1904.03742}

\bibitem{Curiac_2015}
Curiac D~I and Volosencu C 2015 {Imparting protean behavior to mobile robots
  accomplishing patrolling tasks in the presence of adversaries} {\em
  Bioinspir. Biomim.\/} {\bf 10} 056017

\bibitem{Elamvazhuthi_2019}
Elamvazhuthi K and Berman S 2019 Mean-field models in swarm robotics: a survey
  {\em Bioinspir. Biomim.\/} {\bf 15} 015001

\bibitem{20.500.11850/83978}
Alonso-Mora J 2014 {Collaborative motion planning for multi-agent systems} {\em
  Ph.D. thesis\/} (Autonomous Systems Lab, ETH-Z{\"u}rich)

\bibitem{5946136}
{Mohamed} E~F, {El-Metwally} K and {Hanafy} A~R 2011 {An improved Tangent Bug
  method integrated with artificial potential field for multi-robot path
  planning} {\em Int. Symp. on Innovations in Intelligent Systems and
  Applications\/} pp 555--559

\bibitem{baca2020mrs}
B{\'a}{\v{c}}a T, Petrl{\'{i}}k M, Vrba M, Spurn{\'{y}} V, P{\v{e}}ni{\v{c}}ka
  R, He{\v{r}}t D and Saska M 2020 {The MRS UAV System: Pushing the Frontiers
  of Reproducible Research, Real-world Deployment, and Education with
  Autonomous Unmanned Aerial Vehicles} {\bf arXiv:2008.08050}

\bibitem{5717652}
{Lee} T, {Leok} M and {McClamroch} N~H 2010 {Geometric tracking control of a
  quadrotor UAV on SE(3)} {\em 49th Conf. on Decision and Control\/} pp
  5420--5425

\bibitem{8972370}
{Petr{\'a}{\v{c}}ek} P, {Kr{\'a}tk{\'y}} V and {Saska} M 2020 {Dronument:
  System for Reliable Deployment of Micro Aerial Vehicles in Dark Areas of
  Large Historical Monuments} {\em IEEE Trans. Robot. Autom.\/} {\bf 5}
  2078--2085

\bibitem{baca2018mpc}
{B{\'a}{\v{c}}a} T, {Hert} D, {Loianno} G, {Saska} M and {Kumar} V 2018 {Model
  Predictive Trajectory Tracking and Collision Avoidance for Reliable Outdoor
  Deployment of Unmanned Aerial Vehicles} {\em Int. Conf. on Intelligent Robots
  and Systems\/} pp 6753--6760

\bibitem{10.1007/s10514-012-9281-4}
Meier L, Tanskanen P, Heng L, Lee G~H, Fraundorfer F and Pollefeys M 2012
  {PIXHAWK: A Micro Aerial Vehicle Design for Autonomous Flight Using Onboard
  Computer Vision} {\em Auton. Robots\/} {\bf 33} 21--39

\bibitem{1389727}
{Koenig} N and {Howard} A 2004 {Design and use paradigms for Gazebo, an
  open-source multi-robot simulator} {\em Int. Conf. on Intelligent Robots and
  Systems\/} vol~3 pp 2149--2154

\bibitem{ros}
{Stanford Artificial Intelligence Laboratory et al{}} {Robot Operating System}
  \urlprefix\url{ros.org}

\end{thebibliography}

  \end{document}